\documentclass{article}

\usepackage{arxiv}

\usepackage[utf8]{inputenc} 
\usepackage[T1]{fontenc}    
\usepackage{hyperref}       
\usepackage{natbib}         
\usepackage{url}            
\usepackage{booktabs}       
\usepackage{amsfonts}       
\usepackage{nicefrac}       
\usepackage{microtype}      
\usepackage{mathtools}      
\usepackage{lipsum}
\usepackage{graphicx}
\usepackage{amsmath}
\usepackage{amsthm}
\usepackage{amssymb}
\usepackage{algorithm}
\usepackage{algorithmic}
\graphicspath{ {./images/} }
\usepackage{algorithm}
\usepackage{algorithmic}
\usepackage{booktabs}
\usepackage{multirow}
\usepackage{amsthm}
\theoremstyle{plain}
\newtheorem{theorem}{Theorem}[section]

\theoremstyle{definition}

\theoremstyle{remark}

\title{Distributionally Robust Self Paced Curriculum Reinforcement Learning}

\author{
 Anirudh Satheesh \\
  University of Maryland\\
  College Park, MD 20742 \\
  \texttt{anirudhs@terpmail.umd.edu} \\
   \And
 Keenan Powell \\
  University of Maryland\\
  College Park, MD 20742 \\
  \texttt{kpowell1@terpmail.umd.edu} \\
  \And
 Vaneet Aggarwal \\
  Purdue University \\
  West Lafayette, IN 47907 \\
  \texttt{vaneet@purdue.edu} \\
}
 
\begin{document}
\maketitle
\begin{abstract}
A central challenge in reinforcement learning is that policies trained in controlled environments often fail under distribution shifts at deployment into real-world environments. Distributionally Robust Reinforcement Learning (DRRL) addresses this by optimizing for worst-case performance within an uncertainty set defined by a robustness budget $\epsilon$. However, fixing $\epsilon$ results in a tradeoff between performance and robustness: small values yield high nominal performance but weak robustness, while large values can result in instability and overly conservative policies. We propose Distributionally Robust Self-Paced Curriculum Reinforcement Learning (DR-SPCRL), a method that overcomes this limitation by treating $\epsilon$ as a continuous curriculum. DR-SPCRL adaptively schedules the robustness budget according to the agent’s progress, enabling a balance between nominal and robust performance. Empirical results across multiple environments demonstrate that DR-SPCRL not only stabilizes training but also achieves a superior robustness–performance trade-off, yielding an average 24.1\% increase in episodic return under varying perturbations compared to fixed or heuristic scheduling strategies.
\end{abstract}

\section{Introduction}
\label{sec: introduction}
Curriculum Reinforcement Learning (CRL) has emerged as a powerful paradigm for accelerating and stabilizing the training of reinforcement learning (RL) agents. By structuring training from simpler to progressively harder tasks, CRL allows agents to acquire complex skills more reliably and to generalize better to unseen situations \citep{bengio_cl,narvekar2020curriculum}. Instead of exposing an agent to the full complexity of a target task from the beginning, CRL gradually introduces a series of intermediate tasks tailored to the agent’s current level of proficiency. 

While CRL improves learning efficiency and stability, it primarily focuses on generating and learning semantic or goal-specific tasks and, as a result, cannot provide guarantees on performance across arbitrary environment parameterizations. In many real-world applications, policies must contend with unmodeled dynamics, sensor noise, and physical variations between training and deployment, known as the sim-to-real problem. These sources of uncertainty can cause dramatic performance degradation even for policies that performed well during training. To address this, Distributionally Robust Reinforcement Learning (DRRL) \cite{smirnova2019distributionally} provides a principled framework for learning policies that maximize worst-case returns over a prescribed uncertainty set. The resulting policies are explicitly designed to be resilient to model mismatch, in contrast to standard RL approaches that optimize performance only in the nominal environment.

\begin{figure}[htbp]
    \centering
    \includegraphics[width=\linewidth]{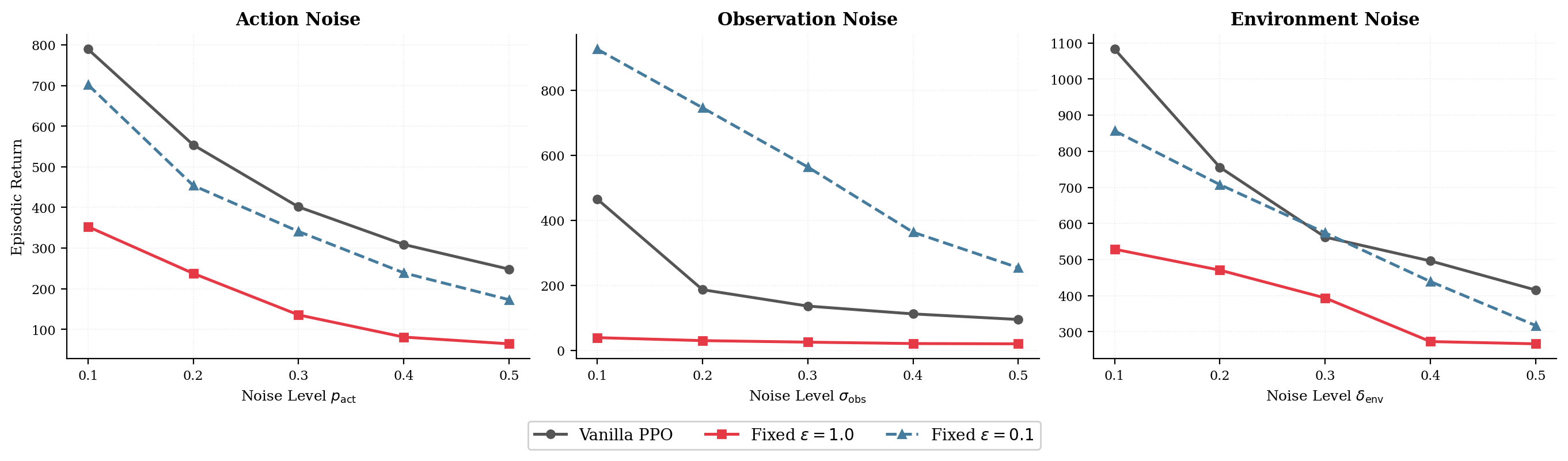}
    \caption{Robustness evaluation of DRPPO under action, observation, and environment perturbations for the Hopper environment. Each panel shows episodic returns as a function of noise levels. Policies trained with smaller fixed robustness budgets ($\epsilon$) fail to handle perturbations effectively, while larger $\epsilon$ values produce overly conservative behavior, leading to suboptimal policies. This highlights the tradeoff between robustness and nominal performance for fixed $\epsilon$ settings.}
    \label{fig:motivation_figure}
\end{figure}

In DRRL, the size of the uncertainty set is determined by a robustness budget \(\epsilon\), which specifies the allowable deviation between the nominal model and its perturbations. As illustrated in Figure~\ref{fig:motivation_figure}, small values of \(\epsilon\) yield high nominal performance but weak robustness, whereas large values guarantee robustness but lead to overly pessimistic value estimates that can slow or destabilize learning. Thus, training policies with a large fixed \(\epsilon\) forces the agent to optimize against a severely depressed value function, while training with small \(\epsilon\) produces policies that are insufficiently robust at deployment. This inherent trade-off motivates the use of curriculum learning in DRRL, where we treat the robustness budget \(\epsilon\) as the context of a curriculum: beginning with manageable uncertainty set and gradually expanding it as the agent’s performance and robustness increases.

\subsection{Challenges and Contributions}

The main challenge of training an effective DRRL policy is to automatically schedule the robustness budget \(\epsilon\) so that an agent can efficiently learn a policy guaranteed to maintain performance under any environment uncertainty. Designing such a schedule is non-trivial: the task difficulty parameterized by \(\epsilon\) lacks an intuitive or direct link to semantic properties of the RL task, and poor scheduling either undermines robustness or slows and destabilizes training. This motivates an approach in which the agent itself determines how quickly to expand the uncertainty set based on its learning progress and robustness level.

To address this problem, we introduce \textbf{Distributionally Robust Self-Paced Curriculum Reinforcement Learning (DR-SPCRL)}, a novel algorithm that automates curriculum generation for \(\epsilon\). We derive DR-SPCRL by instantiating a general curriculum learning framework for the specific mathematical structure of DRRL, with the novelty that the curriculum update is guided directly by the agent's current robustness to environmental uncertainty, enabling it to adaptively balance robustness and performance compared to heuristic-based approaches. By applying the Envelope Theorem to the primal DRRL problem, we formally show that the gradient of the robust value function with respect to the curriculum parameter \(\epsilon\) is equal to the negative of the optimal dual variable, \(\beta^*\) \citep{milgrom2002envelope}. This dual variable represents the marginal cost of robustness, a theoretically grounded measure of how much the agent is struggling at its current robustness level. DR-SPCRL leverages this signal to create an adaptive update rule that automatically balances the agent's competence with the need to progress the curriculum.

In summary, our main contributions are:
\begin{itemize}
    \item We are the first to motivate and formalize the scheduling of the robustness budget \(\epsilon\) in DRRL as a continuous, contextual curriculum learning problem to improve the stability of training DRRL policies.
    \item We introduce DR-SPCRL, a novel automated curriculum algorithm that leverages the dual structure of distributionally robust reinforcement learning (DRRL) to adaptively adjust the robustness budget $\epsilon$ based on the agent’s learning and robustness progress. 
    \item Empirical evaluations across diverse continuous control environments show that DR-SPCRL stabilizes training and achieves superior robustness–performance trade-offs, yielding a $24.1\%$ average improvement in episodic returns compared to both non-robust baselines and DRRL agents trained with fixed or heuristic robustness schedules.
\end{itemize}

\section{Related Work}
\subsection{Curriculum Reinforcement Learning}
Curriculum Learning \citep{bengio_cl} enhances generalization by structuring agent training from simpler to more complex tasks. Early approaches relied on manually ordered tasks or easily parameterized task sets \citep{narvekar2020curriculum, soviany2022curriculumlearningsurvey}. Modern methods automate this process, such as Self-Paced Contextual RL \citep{klink2020self}, which formulates the problem as an active optimization of an intermediate context distribution, balancing local reward maximization with a KL-divergence penalty to ensure progress towards a target distribution. This idea is grounded in information-theoretic frameworks like contextual relative entropy policy search. Other automated methods include using intrinsic motivation \citep{sukhbaatar2018intrinsicmotivationautomaticcurricula}, generative models for the environment \citep{satheesh2025cmalc, florensa2018automaticgoalgenerationreinforcement}, and self-play \citep{du2022takestangomultiagentselfplay}. However, the key limitation of these existing CRL methods is that they are specific to a narrow class of contexts, such as predefined semantic tasks or goal-conditioned settings, and offer no systematic mechanism or guarantees for handling broader or adversarial environmental variations. This motivates our work on Distributionally Robust Self-Paced Curriculum Reinforcement Learning (DR-SPCRL), which explicitly integrates robustness into the curriculum design so that the resulting policies maintain strong performance even under significant deviations from the training conditions.

\subsection{Distributionally Robust Reinforcement Learning}
Distributionally Robust Optimization \citep{kuhn2025distributionallyrobustoptimization,huber1981} refers to optimization in which the underlying parameters are unknown. For Reinforcement Learning \citep{smirnova2019distributionally, liu2022distributionally}, this is formalized as a robust Markov Decision Process \citep{robust_dynamic_programming, wiesmann2013}. Existing literature on DRRL focuses on extending RL methods such as Q-learning \citep{liu2022distributionally, wang2024sample}, value iteration \citep{lu2024distributionally}, and policy gradient methods \citep{wang2022policy, kumar2023policy, zhou2023natural, wang2023policy}. Additional research has focused on the model-based \citep{ramesh2024distributionally} and offline setting \citep{ma2022distributionally, blanchet2023double, liu2024minimax}. However, DRRL is largely limited to the tabular setting \citep{lu2024distributionally, liu2022distributionally, clavier2024}. Recently, some works \citep{cui2025dr} have extended DRRL to the continuous setting by leveraging the closed form of the KL divergence uncertainty set. However, the primary pitfall of existing deep DRRL methods is their reliance on a fixed robustness budget \(\epsilon\). These methods lack an adaptive mechanism to manage this difficulty parameter during training, which can lead to the instability or overly conservative policies that our work aims to resolve.
\section{Problem Formulation}

\subsection{Markov Decision Processes}

We formalize the Distributionally Robust Self-Paced Curriculum Reinforcement Learning (DR-SPCRL) problem by starting from the standard Markov Decision Process (MDP) \(M = (\mathcal{S}, \mathcal{A}, P, r, \gamma, \rho_0)\) where \(\mathcal{S}\) is the state space, \(\mathcal{A}\) is the action space, \(P: \mathcal{S} \times \mathcal{A} \to \Delta(\mathcal{S})\) is the transition kernel (with \(\Delta(\mathcal{S})\) denoting the probability simplex over \(\mathcal{S}\)), \(r: \mathcal{S} \times \mathcal{A} \to \mathbb{R}\) is the bounded reward function, \(\gamma \in [0,1)\) is the discount factor, and \(\rho_0 \in \Delta(\mathcal{S})\) is the initial state distribution. The standard RL objective is to find a policy that maximizes the expected discounted return
\begin{align}
    J(\pi_\theta,P) = \mathbb{E}_{\pi_\theta,P} \left[\sum_{t=0}^{\infty}\gamma^t r(s_t,a_t) \bigg| s_0 \sim \rho_0\right].
\end{align}
The value function and action-value function under policy \(\pi\) are defined as:
\begin{align}
V^{\pi}(s) &= \mathbb{E}_{\pi_\theta, P} \left[\sum_{t=0}^{\infty}\gamma^t r(s_t,a_t)  \Bigg| s_0 = s\right], \\
Q^{\pi}(s,a) &= r(s,a) + \gamma \mathbb{E}_{s' \sim P(\cdot|s,a)}[V^{\pi}(s')].
\end{align}

\subsection{Distributionally Robust MDPs}

In many applications the true transition kernel is not exactly known; instead, we assume it lies in an \emph{uncertainty set} \(\mathcal{P}(\epsilon)\) around a nominal model \(P_0\). For a divergence measure \(D: \Delta(\mathcal{S}) \times \Delta(\mathcal{S}) \to \mathbb{R}_+\), we define:
\begin{align}
    \mathcal{P}(\epsilon) = \bigl\{P: \mathcal{S} \times \mathcal{A} \to \Delta(\mathcal{S}) \,\big|\, D(P(\cdot | s, a), P_0(\cdot | s, a)) \leq \epsilon, \;\forall (s, a) \in \mathcal{S} \times \mathcal{A}\bigr\}
\end{align}
where \(\epsilon\ge0\) is the \emph{robustness budget} controlling the size of the uncertainty set. The resulting robust MDP objective seeks a policy that maximizes the worst-case expected return:
\begin{align}
\max_{\theta}V_{\text{robust}}(\pi_\theta;\epsilon), \qquad V_{\text{robust}}(\pi_\theta;\epsilon) =
\inf_{P\in\mathcal{P}(\epsilon)} J(\pi_\theta,P).
\end{align}
For a fixed policy \(\pi\), the robust value function satisfies the robust Bellman equation:
\begin{align}
    V^{\pi}_{\text{robust}}(s;\epsilon) = \mathbb{E}_{a \sim \pi(\cdot|s)}\!\bigl[Q^{\pi}_{\text{robust}}(s,a;\epsilon)\bigr]
\end{align}
where the robust Q-function is given by:
\begin{equation}
Q^{\pi}_{\text{robust}}(s,a;\epsilon) = r(s,a) + \gamma \inf_{p \in \mathcal{P}_{s,a}(\epsilon)} \mathbb{E}_{s' \sim p}[V^{\pi}_{\text{robust}}(s';\epsilon)]
\end{equation}
and \(\mathcal{P}_{s,a}(\epsilon) = \{p \in \Delta(\mathcal{S}) \mid D(p, P_0(\cdot|s,a)) \leq \epsilon\}\) is the local uncertainty set for state-action pair \((s,a)\).

\subsection{Curriculum Learning for Robust RL}

To model multiple robustness levels within a single framework, we view the robustness budget as a \emph{context variable} \(c\in \mathcal{C}\) with 
\[
\mathcal{C}=[0,\epsilon_{\text{budget}}], 
\qquad
c\equiv\epsilon.
\]
Each context specifies a different uncertainty set \(\mathcal{P}(c)\) and thus a distinct robust control problem, yielding a contextual robust MDP defined by a family of robust MDPs \(\{\mathcal{M}(c)\}_{c \in \mathcal{C}}\). Rather than training directly from the most challenging context \(\epsilon_{\text{budget}}\), distributionally robust curriculum reinforcement learning organizes learning through a curriculum of increasing budgets
\begin{align}
    \mathcal{E}=(\epsilon_1,\epsilon_2,\dots,\epsilon_T), \qquad 0\le\epsilon_1 \le \epsilon_2 \le \cdots \le \epsilon_T \le\epsilon_{\text{budget}},
\end{align}
where each \(\epsilon_{t+1}\) represents a more difficult level of uncertainty than \(\epsilon_t\). At stage \(t\), the agent trains on uncertainty set \(\mathcal{P}(\epsilon_t)\) to learn policy \(\pi_{\theta_t}\):
\begin{align}
    \max_{\theta_t}V_{\text{robust}}(\pi_{\theta_t};\epsilon_t),
\quad t=1,\dots,T,
\end{align}
and the budget is increased only when sufficient progress is achieved. Our objective is therefore to jointly learn a sequence of policies \(\{\pi_{\theta_t}\}_{t=1}^T\) and the curriculum \(\mathcal{E}\) so that the final policy attains strong robust performance at the target uncertainty level \(\epsilon_{\text{budget}}\).
\section{Method}
\label{sec:method}
In this section, we detail our proposed algorithm. We begin by establishing how DRRL can be used in the deep reinforcement learning setting. We then formally derive our curriculum reinforcement learning algorithm, which we term Distributionally Robust Self-Paced Curriculum Reinforcement Learning (DR-SPCRL), by starting with the general curriculum reinforcement learning framework and integrating the distributionally robust formulation.

\subsection{Deep Distributionally Robust RL}
We consider a DRRL setting where the transition model \(P\) is unknown but belongs to a KL-divergence ball \(\mathcal{P}(\epsilon)\) of radius \(\epsilon\) around a nominal model \(P_0\). The goal is to find a policy \(\pi_\theta\) with parameters \(\theta\) that maximizes the robust value function. For a fixed policy \(\pi\), the robust value function is defined by:
\begin{align}
    V^{\pi}_{\text{robust}}(s;\epsilon) = \mathbb{E}_{a \sim \pi(\cdot|s)}\!\left[r(s,a) + \gamma \inf_{p \in \mathcal{P}_{s,a}(\epsilon)} \mathbb{E}_{s' \sim p}[V^{\pi}_{\text{robust}}(s';\epsilon)]\right]
    \label{eq:primal_robust_value}
\end{align}
where \(\mathcal{P}_{s,a}(\epsilon) = \{p \in \Delta(\mathcal{S}) \mid D_{\text{KL}}(p \| P_0(\cdot|s,a)) \leq \epsilon\}\) is the local uncertainty set for state-action pair \((s,a)\). However, solving the infimum in Eq.~\eqref{eq:primal_robust_value} directly is intractable, but strong duality holds for KL-divergence uncertainty sets. Thus, the inner minimization problem is:
\begin{align*}
    \min_{p \in \Delta(\mathcal{S})} \mathbb{E}_{s' \sim p}[V^{\pi}_{\text{robust}}(s';\epsilon)] \quad \text{s.t.} \quad D_{\text{KL}}(p \| P_0(\cdot|s,a)) \leq \epsilon
\end{align*}
We can also write the associated robust Q-function as
\begin{align}
    Q^{\pi}_{\text{robust}}(s,a;\epsilon) = \sup_{\beta > 0} \left\{r(s,a) + \gamma\left( -\beta \log \mathbb{E}_{s' \sim P_0(\cdot | s, a)}\left[\exp\left(-\frac{V^{\pi}_{\text{robust}}(s';\epsilon)}{\beta}\right)\right] - \beta \epsilon \right) \right\}
    \label{eq:robust_q_dual}
\end{align}
where \(\beta > 0\) is the dual variable. In practice, we approximate the optimal dual variable \(\beta^*(s, a)\) using a neural network \(\beta_\phi(s, a)\) with parameters \(\phi\). The agent's policy \(\pi_\theta\) is then trained using any RL algorithm, with policy updates computed using the robust value function \(V^{\pi}_{\text{robust}}\) evaluated via the dual formulation in Eq.~\eqref{eq:robust_q_dual}.

\subsection{The DR-SPCRL Algorithm}

We derive our curriculum generation method from the inference-based self-paced learning framework of \cite{klink2020self}, which poses curriculum generation as a joint optimization problem over the policy parameters \(\theta\) and curriculum parameters \(\nu\):
\begin{align}
\begin{split}
\max_{\theta, \nu} \quad \mathbb{E}_{p(c|\nu)}[J(\theta, c)] - \alpha D_{\mathrm{KL}}(p(c|\nu) \,\|\, \mu(c))
\quad \\ \text{s.t.} \quad 
D_{\mathrm{KL}}(p(c|\nu) \,\|\, p(c|\nu_0)) \le \eta
\label{eq:spcl_objective}
\end{split}
\end{align}
Here, \(J(\theta, c)\) is the performance on context \(c\), \(\mu(c)\) is the target task distribution, \(\alpha\) is a pacing parameter controlling the speed of curriculum progression, and \(\eta\) is a trust region limiting large jumps between contexts. We instantiate this framework for distributionally robust RL by setting \(c = \epsilon\), \(\mu(\epsilon) = \delta(\epsilon - \epsilon_{\text{budget}})\), and \(p(\epsilon|\nu)\) as the curriculum distribution.

We fix the policy \(\theta\) and focus on the curriculum update step. The objective is to find the new curriculum distribution \(p(\epsilon|\nu)\) that solves:
\begin{align}
    \max_{p(\epsilon|\nu)} \quad \int p(\epsilon|\nu) V_{\text{robust}}(\pi_\theta;\epsilon) \, d\epsilon - \alpha D_{\text{KL}}(p(\epsilon|\nu) \| \mu(\epsilon))
    \label{eq:sp_inference_objective}
\end{align}
Because the target distribution \(\mu(\epsilon) = \delta(\epsilon - \epsilon_{\text{budget}})\) is concentrated at a single point, the KL-divergence term strongly incentivizes the optimal curriculum \(p(\epsilon|\nu)\) to also concentrate around a single point. We therefore model the curriculum as a single-point estimate: \(p(\epsilon|\nu) = \delta(\epsilon - \nu)\), where \(\nu\) is the current value of the curriculum's \(\epsilon\). With this simplification, the expectation \(\mathbb{E}_{p(\epsilon|\nu)}[V_{\text{robust}}(\pi_\theta;\epsilon)]\) reduces to \(V_{\text{robust}}(\pi_\theta; \nu)\).

The KL-divergence \(D_{\text{KL}}(\delta(\epsilon - \nu) \| \delta(\epsilon - \epsilon_{\text{budget}}))\) is ill-defined (infinite) between Dirac distributions. To obtain a tractable objective, we approximate both Dirac distributions as narrow isotropic Gaussians: \(\delta(\epsilon - \nu) \approx \mathcal{N}(\nu, \sigma^2)\) and \(\delta(\epsilon - \epsilon_{\text{budget}}) \approx \mathcal{N}(\epsilon_{\text{budget}}, \sigma^2)\) with small variance \(\sigma^2\). The KL-divergence between these Gaussians is:
\begin{align}
    D_{\text{KL}}(\mathcal{N}(\nu, \sigma^2) \| \mathcal{N}(\epsilon_{\text{budget}}, \sigma^2)) = \frac{(\nu - \epsilon_{\text{budget}})^2}{2\sigma^2}
\end{align}
Absorbing the constant \(\frac{1}{2\sigma^2}\) into the pacing parameter \(\alpha\), we obtain a squared Euclidean penalty term \(\alpha(\nu - \epsilon_{\text{budget}})^2\). By the Dirac delta definition and our use of the pointwise parameterization, we can replace \(\nu\) with \(\epsilon\). This transforms the distributional optimization problem into an equivalent optimization over the point-wise curriculum parameter, reducing our curriculum learning problem to
\begin{align}
    \max_{\epsilon} \quad V_{\text{robust}}(\pi_\theta; \epsilon) - \alpha (\epsilon - \epsilon_{\text{budget}})^2 \quad \text{s.t.} \quad (\epsilon - \epsilon_t)^2 \leq \eta
    \label{eq:constrained_curriculum_optimization}
\end{align}
where the constraint ensures the curriculum does not jump too far from the current budget \(\epsilon_t\). To solve this, let \(f(\epsilon; \theta) = V_{\text{robust}}(\pi_\theta; \epsilon) - \alpha (\epsilon - \epsilon_{\text{budget}})^2\) be our objective. The Lagrangian for this constrained optimization is:
\begin{align}
    \mathcal{L}_{\text{curr}}(\epsilon, \lambda) = f(\epsilon; \theta) - \lambda \left( (\epsilon - \epsilon_t)^2 - \eta \right) 
    \label{eq:curriculum_lagrangian}
\end{align}
where \(\lambda \geq 0\) is the Lagrange multiplier for the trust-region constraint. The first-order optimality condition is:
\begin{align}
    \frac{\partial \mathcal{L}_{\text{curr}}}{\partial \epsilon} &= \frac{\partial f(\epsilon; \theta)}{\partial \epsilon} - 2\lambda(\epsilon - \epsilon_t) = \frac{\partial V_{\text{robust}}(\pi_\theta; \epsilon)}{\partial \epsilon} - 2\alpha(\epsilon - \epsilon_{\text{budget}}) - 2\lambda(\epsilon - \epsilon_t) \stackrel{\mathrm{set}}{=} 0
    \label{eq:stationary_condition}
\end{align}
The key challenge is computing \(\frac{\partial V_{\text{robust}}(\pi_\theta; \epsilon)}{\partial \epsilon}\). Recall that \(V_{\text{robust}}(\pi_\theta; \epsilon)\) is defined implicitly as a fixed point of the robust Bellman operator. We leverage the Envelope Theorem \citep{milgrom2002envelope} to differentiate the operator. Let the objective inside the supremum of the robust Bellman operator be \(G(\beta, \epsilon, V')\). The Envelope Theorem states that the derivative with respect to \(\epsilon\) is the partial derivative of the objective evaluated at the optimum \(\beta^*\):
\begin{align}
    \frac{\partial Q^{\pi}_{\text{robust}}(s,a;\epsilon)}{\partial \epsilon} &= \gamma \left( \left. \frac{\partial G}{\partial \epsilon} \right|_{\beta^*} + \mathbb{E}_{s' \sim P^*_\epsilon} \left[ \frac{\partial V^{\pi}_{\text{robust}}(s';\epsilon)}{\partial \epsilon} \right] \right) \\
    &= \gamma \left( -\beta^*(s, a; \epsilon) + \mathbb{E}_{s' \sim P^*_\epsilon} \left[ \frac{\partial V^{\pi}_{\text{robust}}(s';\epsilon)}{\partial \epsilon} \right] \right)
\end{align}
Here, the explicit derivative with respect to \(\epsilon\) yields \(-\beta^*\), while the dependence via the next-state value function produces the expectation under the worst-case transition model \(P^*_\epsilon\). Unrolling this recursive definition over the infinite horizon, we obtain the total sensitivity as the expected discounted sum of the dual variables:
\begin{align}
    \frac{\partial V_{\text{robust}}(\pi_\theta; \epsilon)}{\partial \epsilon} = -\mathbb{E}_{\tau \sim (\pi, P^*_\epsilon)} \left[ \sum_{t=0}^\infty \gamma^{t+1} \beta^*(s_t, a_t; \epsilon) \right]
\end{align}
Using the discounted state-action occupancy measure \(d^{\pi_{\theta}}_{P^*_\epsilon}\), this simplifies to:
\begin{align}
    \frac{\partial V_{\text{robust}}(\pi_\theta; \epsilon)}{\partial \epsilon} = - \frac{\gamma}{1-\gamma} \mathbb{E}_{(s,a) \sim d^{\pi_{\theta}}_{P^*_\epsilon}}\left[\beta^*(s,a; \epsilon)\right]
    \label{eq:envelope_result}
\end{align}
This formally shows that the gradient of the robust value function involves the expected optimal dual variable scaled by the effective horizon. Substituting Eq.~\eqref{eq:envelope_result} into Eq.~\eqref{eq:stationary_condition} yields:
\begin{align} 
    - \frac{\gamma}{1-\gamma} \mathbb{E}_{(s,a) \sim d^{\pi_{\theta}}_{P^*_\epsilon}}\left[\beta^*(s,a; \epsilon)\right] - 2\alpha(\epsilon - \epsilon_{\text{budget}}) - 2\lambda(\epsilon - \epsilon_t) = 0
    \label{eq:optimality_with_beta}
\end{align}
Solving for \(\epsilon\):
\begin{align}
    \epsilon = \frac{2\alpha \epsilon_{\text{budget}} + 2\lambda \epsilon_t - \frac{\gamma}{1-\gamma} \mathbb{E}\left[\beta^*(s,a; \epsilon)\right]}{2(\alpha + \lambda)}
\end{align}
To obtain a practical explicit update rule, we modify this fixed point equation as one Picard iteration, treating the step-size term \(\frac{1}{2(\alpha + \lambda)}\) as an effective curriculum learning rate, denoted \(\lambda_{\text{curr}}\), and defining \(C_\gamma = \frac{\gamma}{1-\gamma}\):
\begin{align}
    \epsilon_{t+1} = \epsilon_t - \lambda_{\text{curr}} \left( C_\gamma \mathbb{E}_{(s,a) \sim d^{\pi_\theta}_{P_0}}[\beta^*(s,a; \epsilon_t)] + 2\alpha(\epsilon_t - \epsilon_{\text{budget}})\right)
    \label{eq:curriculum_update_rule}
\end{align}

\subsection{Practical Implementation}

In practice, we do not have access to \(\beta^*(s,a)\), so we use \(\beta_\phi(s,a)\) as an approximation. We estimate the expected dual variable using a minibatch \(\mathcal{B}\) from the replay buffer. Our final algorithm is RL-algorithm-agnostic and proceeds in a block-coordinate ascent fashion. We provide the full algorithm in Algorithm~\ref{alg:dr-spcl}.

\begin{algorithm}[htbp]
\caption{Distributionally Robust Self-Paced Curriculum Reinforcement Learning (DR-SPCRL)}
\label{alg:dr-spcl}
\begin{algorithmic}[1]
\STATE \textbf{Require:} discount factor \(\gamma\), initial budget \(\epsilon_{\text{start}}\), final budget \(\epsilon_{\text{budget}}\), pacing parameter \(\alpha\), curriculum learning rate \(\lambda_{\text{curr}}\), iterations \(T\), rollouts \(N\), critic updates \(K\).
\STATE Initialize policy \(\pi_\theta\), dual model \(\beta_\phi\), curriculum \(\epsilon_t \leftarrow \epsilon_{\text{start}}\), buffer \(\mathcal{D}\).
\STATE Set \(C_\gamma \gets \frac{\gamma}{1-\gamma}\).
\FOR{\(t = 1\) to \(T\)}
    \STATE Collect \(N\) steps of experience \((s, a, r, s')\) using \(\pi_\theta\); store in \(\mathcal{D}\).
    \FOR{\(k = 1\) to \(K\)}
        \STATE Sample minibatch \(\{(s_j, a_j, r_j, s'_j)\}\) from \(\mathcal{D}\).
        \STATE Compute dual loss \(L(\phi) = -V_{\text{robust}}(\pi_\theta; \epsilon_t)\) using Eq.~\eqref{eq:robust_q_dual}.
        \STATE Update \(\beta_\phi\) via gradient ascent to maximize \(V_{\text{robust}}\).
    \ENDFOR
    \STATE Compute robust value/advantages using updated \(\beta_\phi\) and Eq.~\eqref{eq:robust_q_dual}.
    \STATE Update policy parameters \(\theta\) via RL algorithm.
    \STATE Estimate \(\mathbb{E}[\beta^*] \approx \frac{1}{|\mathcal{B}|}\sum_{(s,a) \in \mathcal{B}} \beta_\phi(s,a)\) from minibatch \(\mathcal{B}\).
    \STATE Compute regularization gradient \(g_{\text{reg}} = 2\alpha(\epsilon_t - \epsilon_{\text{budget}})\).
    \STATE Update curriculum:
    \(\epsilon_{t+1} \leftarrow \epsilon_t - \lambda_{\text{curr}} \left( C_\gamma \frac{1}{|\mathcal{B}|}\sum_{(s,a) \in \mathcal{B}} \beta_\phi(s,a) + g_{\text{reg}}\right)\)
    \STATE Project \(\epsilon_{t+1}\) into \([0, \epsilon_{\text{budget}}]\).
\ENDFOR
\RETURN \(\pi_\theta\).
\end{algorithmic}
\end{algorithm}
\section{Theoretical Analysis}
\label{sec:theory}

We analyze DR-SPCRL as stochastic block-coordinate ascent on the joint objective
\[
F(\theta,\epsilon) = J(\pi_\theta,\epsilon) - \alpha(\epsilon-\epsilon_{\text{budget}})^2,
\]
where $J(\pi_\theta,\epsilon)$ denotes the robust return under curriculum budget $\epsilon$. The algorithm alternates between a stochastic policy-gradient update in $\theta$ and a dual-gradient update in $\epsilon$. We impose standard regularity conditions: $L$-smoothness of $F$, bounded policy gradients, bounded stochastic gradient variance $\sigma^2$, and Lipschitz continuity of the dual response $\beta^*(\epsilon)$ with constant $L_\beta$.

\begin{theorem}[Finite-Time Convergence and Performance Bounds]
\label{thm:convergence}
Let Algorithm~\ref{alg:dr-spcl} be run for $T \ge 4L^2$ iterations with step size $\eta=T^{-1/2}$. Then, assuming the stated regularity conditions, the last iterate $(\pi_T, \epsilon_T)$ satisfies the following bounds:

\begin{enumerate}
    \item \textbf{Policy Stationarity at budget:} The policy $\pi_T$ is an approximate stationary point of the optimal robust objective $J(\cdot, \epsilon_{\mathrm{budget}})$, with expected gradient norm bounded by:
    \[
    \mathbb{E}\|\nabla_\theta J(\pi_T, \epsilon_{\mathrm{budget}})\|
    \;\le\; \mathcal{O}(T^{-1/4}) \;+\; \frac{\gamma L_\beta\,\mathbb{E}[\beta_T^*]}{2\alpha (1-\gamma)}.
    \]

    \item \textbf{Curriculum Approximation Gap:} The expected discrepancy between the robust value at \(\epsilon_{\mathrm{budget}}\) and \(\epsilon_T\) is bounded by:
    \[
    \mathbb{E}\big| J(\pi_T, \epsilon_{\mathrm{budget}}) - J(\pi_T, \epsilon_T) \big|
    \;\le\; \mathcal{O}(T^{-1/2}) \;+\; \frac{\gamma^2(\mathbb{E}[\beta_T^*])^2}{2\alpha(1-\gamma)^2}\left(1 + \frac{L_\beta}{2\alpha}\right).
    \]
\end{enumerate}
Here, $\beta_T^*$ denotes the optimal dual variable for the uncertainty set at $(\pi_T, \epsilon_T)$, and the constants hidden in $\mathcal{O}(\cdot)$ depend on $L$, $\sigma^2$, and the initial function value gap $F(z_0) - F_{\inf}$.
\end{theorem}

We provide the full proof in Section~\ref{appendix: theoretical analysis}. We analyze Algorithm~\ref{alg:dr-spcl} as a stochastic block-coordinate ascent on the joint potential function $F(\theta, \epsilon) = J(\pi_\theta, \epsilon) - \alpha(\epsilon - \epsilon_{\text{budget}})^2$. Leveraging the $L$-smoothness of the objective and the bounded variance of the gradient estimators, we invoke standard non-convex optimization results to establish that the algorithm reaches an $\mathcal{O}(T^{-1/4})$-approximate stationary point in expectation. We then decompose the total performance gap $|J(\pi_T, \epsilon_{\text{budget}}) - J(\pi_T, \epsilon_T)|$. Using the Lipschitz continuity of the dual gradient, we show that it has an error floor that scales quadratically with marginal cost of robustness and inversely with the pacing parameter $\alpha$.
\section{Experiments}
We conduct a comprehensive set of experiments to empirically validate our proposed Distributionally Robust Self-Paced Curriculum Reinforcement Learning (DR-SPCRL) algorithm. Our evaluation targets three key hypotheses: (1) DR-SPCRL produces policies with greater robustness than standard non-robust baselines; (2) its adaptive curriculum yields more stable training and better final performance than robust training with a fixed or heuristic schedule; and (3) the method is general-purpose, providing benefits across diverse algorithms and domains. To demonstrate this, we integrate DR-SPCRL with three state-of-the-art deep RL algorithms representing distinct paradigms: Proximal Policy Optimization (PPO) \citep{schulman2017proximal}, Soft Actor-Critic \citep{haarnoja2018soft} (SAC), and Deep Deterministic Policy Gradients (DDPG) \citep{lillicrap2015continuous}.

We evaluate on Hopper, Humanoid, Half-Cheetah, and Walker2d. Performance is measured using the robust return, defined as the average episodic return under held-out perturbations not encountered during training. Robustness is evaluated across three perturbation types modeling sensor noise, actuator noise, and environment shifts.

\paragraph{Observation noise.}
At each timestep, the agent observes a perturbed state
\begin{align}
s'_t = s_t + \eta_t, \quad \eta_t \sim \mathcal{N}(0,\sigma_{\text{obs}}^2 I),
\end{align}
where Gaussian noise is applied independently to each state dimension.

\paragraph{Action noise.}
With probability \(p_{\text{act}}\), the policy action is replaced by a random action sampled uniformly from the action space:
\begin{align}
a'_t =
\begin{cases}
a_{\text{rand}} \sim U(\mathcal{A}) & p_{\text{act}},\\
\pi(s_t) & 1-p_{\text{act}}.
\end{cases}
\end{align}
\paragraph{Environment noise.}
To model sim-to-real gaps, MuJoCo physics parameters are perturbed at the start of each episode. Given nominal parameters \(\xi_0\), each component is independently rescaled:
\begin{align}
\xi'_i \sim U(1-\delta_{\text{env}},\,1+\delta_{\text{env}})\,\xi_{0,i},
\end{align}
and the episode is executed entirely under the perturbed dynamics \(\xi'\).
 We compare our DR-SPCRL algorithm to six curriculum reinforcement learning algorithms where we set the context as the epsilon radius of the uncertainty set:
\begin{itemize}
    \item \textbf{Vanilla} The standard, non-robust RL algorithm, which corresponds to training with $\epsilon=0$. This serves as our baseline for nominal performance and demonstrates the fragility of standard policies.
    \item \textbf{Fixed Budget} Distributionally Robust RL trained with a fixed, high robustness budget throughout training. This baseline is designed to highlight the instability or over-conservatism that arises from training without a curriculum.
    \item \textbf{Linear Schedule} A baseline that employs a handcrafted curriculum where the robustness budget $\epsilon$ is linearly annealed from 0 to the final target budget over the course of the training run.
    \item \textbf{Domain Randomization} Robust RL trained with a randomly sampled \(\epsilon\) in \([0, \epsilon_{\mathrm{budget}}]\) \citep{tobin2017domain}.
    \item \textbf{ACCEL} An adaptation of the regret-based curriculum generation method, ACCEL \citep{parker2022evolving}. We maintain a replay buffer of $\epsilon$ values, where the regret score of an $\epsilon$ is defined by the mean dual variable $\beta^*$. The curriculum progresses by sampling and editing high-regret $\epsilon$ values from this buffer.
    \item \textbf{SPACE} An adaptation of the Self-Paced Contextual Evaluation (SPACE) algorithm \citep{eimer2021self}. This method employs a heuristic where the curriculum budget $\epsilon$ is increased only when the agent's robust value function has plateaued, indicating that the agent has mastered the current difficulty level.
\end{itemize}
For all experiments we set the learning rate for the \(\phi\) network to be \(5 \times 10^{-4}\) and the number of steps \(K\) to be \(5\). We also set the \(\epsilon_{\text{budget}}\) to be 1, which under the KL divergence captures a wide range of possible transition kernel perturbations. We evaluate each policy after training for 1 million steps for 100 episodes, and report the mean and the standard error for a 95\% confidence interval over 5 seeds. All experiments were conducted on an RTX 4080 Laptop GPU with 4 i9-13900HX CPUs.

\section{Results}
\label{sec: results}
\begin{table*}[t]
\centering
\resizebox{\textwidth}{!}{%
\begin{tabular}{llccccccccccccccc}
\toprule
\textbf{Env.} & \textbf{Alg.} & \multicolumn{5}{c}{\textbf{Action}} & \multicolumn{5}{c}{\textbf{Observation}} & \multicolumn{5}{c}{\textbf{Environment}} \\
\cmidrule(lr){3-7} \cmidrule(lr){8-12} \cmidrule(lr){13-17}
 &  & $p_{\text{act}}=0.1$ & $p_{\text{act}}=0.2$ & $p_{\text{act}}=0.3$ & $p_{\text{act}}=0.4$ & $p_{\text{act}}=0.5$ & $\sigma_{\text{obs}}=0.1$ & $\sigma_{\text{obs}}=0.2$ & $\sigma_{\text{obs}}=0.3$ & $\sigma_{\text{obs}}=0.4$ & $\sigma_{\text{obs}}=0.5$ & $\delta_{\text{env}}=0.1$ & $\delta_{\text{env}}=0.2$ & $\delta_{\text{env}}=0.3$ & $\delta_{\text{env}}=0.4$ & $\delta_{\text{env}}=0.5$ \\
\midrule
\multirow{7}{*}{\texttt{HalfCheetah}} & PPO & $1168.0_{(271.2)}$ & $\underline{869.3}_{(163.7)}$ & $\underline{633.1}_{(123.4)}$ & $\underline{424.3}_{(78.4)}$ & $\underline{249.5}_{(47.9)}$ & $1279.6_{(527.6)}$ & $882.5_{(468.5)}$ & $566.0_{(420.8)}$ & $337.9_{(344.3)}$ & $175.0_{(284.1)}$ & $1467.6_{(490.6)}$ & $1222.8_{(446.3)}$ & $855.7_{(325.6)}$ & $602.3_{(311.8)}$ & $385.5_{(227.2)}$ \\
 & Fixed & $1158.1_{(516.7)}$ & $760.7_{(333.5)}$ & $513.0_{(297.7)}$ & $246.5_{(161.9)}$ & $105.3_{(131.9)}$ & $1040.5_{(316.3)}$ & $633.7_{(170.6)}$ & $321.3_{(148.3)}$ & $162.0_{(158.4)}$ & $34.0_{(128.8)}$ & $1516.9_{(706.2)}$ & $1262.7_{(717.8)}$ & $997.7_{(621.0)}$ & $709.4_{(549.9)}$ & $\underline{477.9}_{(434.9)}$ \\
 & SPACE & $1005.8_{(58.8)}$ & $759.7_{(58.6)}$ & $541.9_{(55.1)}$ & $349.4_{(49.6)}$ & $187.0_{(43.5)}$ & $960.8_{(29.1)}$ & $541.6_{(75.3)}$ & $235.1_{(73.4)}$ & $52.3_{(62.0)}$ & $-53.0_{(46.6)}$ & $1218.6_{(62.4)}$ & $979.4_{(76.2)}$ & $665.4_{(47.4)}$ & $366.9_{(53.1)}$ & $180.0_{(61.2)}$ \\
 & Linear & $985.2_{(20.2)}$ & $753.2_{(29.9)}$ & $535.7_{(34.9)}$ & $356.9_{(32.4)}$ & $201.1_{(23.8)}$ & $945.9_{(22.7)}$ & $522.9_{(35.9)}$ & $218.1_{(45.9)}$ & $27.5_{(38.6)}$ & $-82.0_{(39.4)}$ & $1167.4_{(39.4)}$ & $921.4_{(9.6)}$ & $617.4_{(56.2)}$ & $367.8_{(25.9)}$ & $133.7_{(45.2)}$ \\
 & ACCEL & $1110.5_{(359.0)}$ & $782.0_{(225.3)}$ & $520.9_{(173.9)}$ & $319.4_{(147.6)}$ & $160.0_{(132.8)}$ & $1263.0_{(436.3)}$ & $867.6_{(395.5)}$ & $509.1_{(336.6)}$ & $279.0_{(291.8)}$ & $117.6_{(241.4)}$ & $1448.3_{(518.4)}$ & $1213.8_{(443.9)}$ & $903.7_{(403.9)}$ & $636.7_{(402.7)}$ & $347.4_{(342.6)}$ \\
 & Dom. Rand. & $\underline{1243.1}_{(416.0)}$ & $843.1_{(235.4)}$ & $527.6_{(147.2)}$ & $297.7_{(107.7)}$ & $114.1_{(60.2)}$ & $\underline{1396.7}_{(385.2)}$ & $\underline{997.0}_{(354.6)}$ & $\underline{579.9}_{(317.5)}$ & $\underline{395.8}_{(260.2)}$ & $\underline{242.3}_{(214.5)}$ & $\underline{1717.0}_{(772.8)}$ & $\underline{1385.8}_{(583.0)}$ & $\underline{1082.6}_{(559.6)}$ & $\underline{722.5}_{(470.9)}$ & $448.4_{(344.8)}$ \\
 & Self-Paced & $\mathbf{1834.0}_{(207.5)}$ & $\mathbf{1182.1}_{(63.3)}$ & $\mathbf{800.6}_{(57.2)}$ & $\mathbf{507.7}_{(52.2)}$ & $\mathbf{266.3}_{(33.4)}$ & $\mathbf{2184.5}_{(139.2)}$ & $\mathbf{1570.3}_{(201.9)}$ & $\mathbf{1131.1}_{(176.9)}$ & $\mathbf{761.4}_{(147.7)}$ & $\mathbf{545.5}_{(110.0)}$ & $\mathbf{2338.5}_{(564.8)}$ & $\mathbf{2017.2}_{(337.6)}$ & $\mathbf{1601.8}_{(207.0)}$ & $\mathbf{1215.1}_{(82.3)}$ & $\mathbf{935.4}_{(146.1)}$ \\
\midrule
\multirow{7}{*}{\texttt{Walker2d}} & PPO & $\underline{1048.3}_{(435.5)}$ & $651.4_{(202.5)}$ & $453.2_{(65.8)}$ & $349.7_{(30.8)}$ & $257.7_{(23.9)}$ & $732.4_{(229.0)}$ & $392.0_{(43.2)}$ & $\underline{333.6}_{(26.9)}$ & $260.3_{(25.4)}$ & $208.7_{(28.7)}$ & $\underline{1624.5}_{(712.4)}$ & $\underline{1199.6}_{(446.5)}$ & $\underline{914.8}_{(382.2)}$ & $\underline{714.9}_{(184.4)}$ & $\underline{610.7}_{(159.5)}$ \\
 & Fixed & $425.8_{(77.8)}$ & $340.9_{(54.5)}$ & $263.6_{(76.2)}$ & $166.8_{(83.0)}$ & $100.6_{(84.2)}$ & $365.2_{(54.5)}$ & $176.2_{(91.1)}$ & $95.6_{(92.6)}$ & $62.3_{(68.4)}$ & $44.0_{(54.4)}$ & $477.1_{(91.3)}$ & $443.9_{(93.2)}$ & $393.4_{(69.9)}$ & $338.5_{(79.4)}$ & $293.1_{(71.9)}$ \\
 & SPACE & $1020.1_{(277.6)}$ & $\mathbf{708.9}_{(156.9)}$ & $\mathbf{494.6}_{(57.4)}$ & $\underline{370.6}_{(21.8)}$ & $264.1_{(20.4)}$ & $\underline{823.3}_{(264.1)}$ & $\underline{425.0}_{(65.4)}$ & $308.4_{(37.7)}$ & $241.8_{(38.1)}$ & $195.5_{(41.2)}$ & $1352.0_{(444.5)}$ & $1068.0_{(279.0)}$ & $908.4_{(196.8)}$ & $714.0_{(137.9)}$ & $580.5_{(59.6)}$ \\
 & Linear & $667.4_{(54.0)}$ & $515.8_{(44.5)}$ & $427.6_{(43.0)}$ & $353.2_{(29.1)}$ & $\mathbf{293.0}_{(16.9)}$ & $544.6_{(75.3)}$ & $347.3_{(19.1)}$ & $255.5_{(30.9)}$ & $193.4_{(35.4)}$ & $146.1_{(31.5)}$ & $895.8_{(114.6)}$ & $734.0_{(92.3)}$ & $637.3_{(114.8)}$ & $536.4_{(49.4)}$ & $512.3_{(64.2)}$ \\
 & ACCEL & $616.6_{(162.7)}$ & $441.7_{(48.2)}$ & $361.2_{(30.6)}$ & $313.0_{(24.0)}$ & $232.9_{(33.3)}$ & $501.0_{(162.0)}$ & $382.1_{(89.3)}$ & $307.9_{(31.0)}$ & $\underline{272.4}_{(17.0)}$ & $\underline{235.5}_{(33.5)}$ & $1021.3_{(522.2)}$ & $812.4_{(325.1)}$ & $652.9_{(198.9)}$ & $534.7_{(111.2)}$ & $461.0_{(73.0)}$ \\
 & Dom. Rand. & $473.8_{(154.9)}$ & $366.4_{(75.5)}$ & $288.4_{(86.6)}$ & $224.2_{(111.7)}$ & $156.4_{(90.6)}$ & $339.8_{(43.0)}$ & $249.4_{(84.7)}$ & $211.1_{(103.1)}$ & $174.1_{(96.7)}$ & $129.3_{(81.4)}$ & $642.0_{(358.6)}$ & $561.3_{(272.5)}$ & $502.8_{(225.9)}$ & $408.8_{(96.2)}$ & $367.7_{(68.4)}$ \\
 & Self-Paced & $\mathbf{1084.8}_{(164.5)}$ & $\underline{696.5}_{(72.3)}$ & $\underline{478.9}_{(22.1)}$ & $\mathbf{378.0}_{(21.3)}$ & $\underline{282.4}_{(9.7)}$ & $\mathbf{1070.5}_{(366.5)}$ & $\mathbf{581.3}_{(202.5)}$ & $\mathbf{441.1}_{(96.5)}$ & $\mathbf{371.2}_{(63.3)}$ & $\mathbf{290.9}_{(68.7)}$ & $\mathbf{2014.5}_{(277.0)}$ & $\mathbf{1439.5}_{(248.5)}$ & $\mathbf{1077.9}_{(121.1)}$ & $\mathbf{870.3}_{(73.7)}$ & $\mathbf{635.9}_{(49.1)}$ \\
\midrule
\multirow{7}{*}{\texttt{Humanoid}} & PPO & $588.0_{(22.6)}$ & $554.2_{(26.5)}$ & $516.4_{(22.2)}$ & $455.4_{(20.9)}$ & $384.4_{(18.3)}$ & $208.3_{(10.4)}$ & $207.3_{(8.7)}$ & $200.8_{(12.6)}$ & $203.9_{(8.4)}$ & $203.3_{(7.3)}$ & $133.2_{(11.8)}$ & $134.9_{(12.2)}$ & $131.7_{(6.2)}$ & $128.2_{(4.6)}$ & $126.5_{(9.2)}$ \\
 & Fixed & $499.8_{(31.6)}$ & $439.4_{(34.8)}$ & $336.7_{(38.4)}$ & $242.8_{(23.2)}$ & $176.7_{(13.3)}$ & $133.5_{(2.3)}$ & $134.4_{(5.2)}$ & $140.6_{(5.7)}$ & $130.6_{(5.3)}$ & $136.6_{(5.7)}$ & $132.1_{(14.0)}$ & $131.2_{(10.6)}$ & $136.8_{(10.4)}$ & $135.2_{(11.0)}$ & $127.6_{(10.6)}$ \\
 & SPACE & $598.5_{(44.3)}$ & $566.7_{(39.7)}$ & $516.5_{(26.5)}$ & $452.0_{(15.1)}$ & $\underline{386.0}_{(14.5)}$ & $214.3_{(9.4)}$ & $216.4_{(13.2)}$ & $207.2_{(12.6)}$ & $214.4_{(11.6)}$ & $210.4_{(12.9)}$ & $137.5_{(5.7)}$ & $131.4_{(13.3)}$ & $134.7_{(7.1)}$ & $138.3_{(7.9)}$ & $130.6_{(10.8)}$ \\
 & Linear & $599.0_{(52.3)}$ & $574.0_{(50.7)}$ & $508.2_{(39.7)}$ & $\underline{456.2}_{(28.6)}$ & $372.0_{(24.4)}$ & $228.7_{(12.1)}$ & $227.4_{(10.6)}$ & $229.3_{(11.8)}$ & $\underline{236.1}_{(6.4)}$ & $234.0_{(8.6)}$ & $146.5_{(11.3)}$ & $\underline{143.1}_{(9.1)}$ & $146.2_{(12.0)}$ & $139.4_{(6.6)}$ & $136.9_{(11.6)}$ \\
 & ACCEL & $\underline{622.9}_{(105.7)}$ & $\underline{577.3}_{(66.1)}$ & $\underline{529.0}_{(52.3)}$ & $451.9_{(27.2)}$ & $359.4_{(13.2)}$ & $\underline{235.6}_{(11.6)}$ & $\underline{228.8}_{(16.7)}$ & $\underline{233.1}_{(21.1)}$ & $233.5_{(17.3)}$ & $\underline{235.4}_{(10.3)}$ & $\underline{158.4}_{(18.5)}$ & $137.7_{(13.1)}$ & $\underline{150.3}_{(10.6)}$ & $\underline{143.6}_{(16.4)}$ & $\underline{137.7}_{(11.8)}$ \\
 & Dom. Rand. & $542.1_{(34.4)}$ & $508.0_{(43.1)}$ & $411.4_{(30.3)}$ & $325.0_{(45.5)}$ & $237.6_{(41.4)}$ & $151.4_{(15.4)}$ & $159.4_{(11.7)}$ & $163.7_{(15.3)}$ & $153.3_{(15.1)}$ & $153.4_{(9.8)}$ & $143.7_{(10.2)}$ & $138.0_{(18.8)}$ & $129.8_{(12.3)}$ & $133.0_{(8.8)}$ & $131.9_{(10.0)}$ \\
 & Self-Paced & $\mathbf{680.1}_{(51.0)}$ & $\mathbf{633.6}_{(28.6)}$ & $\mathbf{571.0}_{(9.6)}$ & $\mathbf{469.4}_{(12.4)}$ & $\mathbf{399.5}_{(12.6)}$ & $\mathbf{332.6}_{(168.4)}$ & $\mathbf{331.6}_{(170.6)}$ & $\mathbf{325.6}_{(155.1)}$ & $\mathbf{307.0}_{(148.0)}$ & $\mathbf{317.5}_{(147.9)}$ & $\mathbf{215.8}_{(98.7)}$ & $\mathbf{206.9}_{(104.8)}$ & $\mathbf{218.8}_{(116.5)}$ & $\mathbf{213.1}_{(129.7)}$ & $\mathbf{215.1}_{(113.4)}$ \\
\midrule
\multirow{7}{*}{\texttt{Hopper}} & PPO & $789.5_{(49.0)}$ & $553.9_{(35.3)}$ & $401.4_{(69.5)}$ & $308.9_{(67.6)}$ & $248.3_{(54.8)}$ & $465.3_{(232.5)}$ & $186.8_{(121.0)}$ & $136.4_{(59.8)}$ & $112.4_{(31.1)}$ & $95.2_{(17.7)}$ & $\underline{1082.6}_{(122.3)}$ & $\underline{755.6}_{(74.5)}$ & $562.2_{(14.9)}$ & $\underline{496.2}_{(39.6)}$ & $415.7_{(48.9)}$ \\
 & Fixed & $352.8_{(329.1)}$ & $237.7_{(229.8)}$ & $135.6_{(105.3)}$ & $81.1_{(72.2)}$ & $64.5_{(58.5)}$ & $39.4_{(30.6)}$ & $30.2_{(22.0)}$ & $25.5_{(15.4)}$ & $21.2_{(10.2)}$ & $20.3_{(9.5)}$ & $528.6_{(382.8)}$ & $470.6_{(322.1)}$ & $393.6_{(252.2)}$ & $272.8_{(195.1)}$ & $266.5_{(214.3)}$ \\
 & SPACE & $\underline{795.5}_{(79.8)}$ & $\underline{589.8}_{(55.5)}$ & $\underline{428.1}_{(37.3)}$ & $\underline{343.6}_{(27.2)}$ & $\underline{276.0}_{(35.1)}$ & $\underline{556.7}_{(212.2)}$ & $\underline{223.9}_{(89.7)}$ & $\underline{157.7}_{(48.9)}$ & $\underline{120.2}_{(24.9)}$ & $\underline{107.1}_{(22.5)}$ & $1012.3_{(110.8)}$ & $726.3_{(70.5)}$ & $566.0_{(34.1)}$ & $480.2_{(21.8)}$ & $\underline{447.0}_{(43.4)}$ \\
 & Linear & $716.5_{(219.4)}$ & $479.0_{(123.0)}$ & $346.1_{(82.9)}$ & $237.4_{(48.9)}$ & $181.6_{(29.9)}$ & $156.5_{(79.3)}$ & $96.4_{(19.1)}$ & $93.9_{(12.1)}$ & $93.2_{(8.9)}$ & $94.1_{(7.7)}$ & $894.1_{(307.7)}$ & $702.3_{(178.3)}$ & $\underline{585.0}_{(133.1)}$ & $485.6_{(94.2)}$ & $412.8_{(64.0)}$ \\
 & ACCEL & $451.2_{(234.2)}$ & $358.9_{(123.8)}$ & $302.0_{(72.6)}$ & $260.7_{(61.1)}$ & $229.0_{(44.2)}$ & $275.5_{(99.1)}$ & $180.0_{(66.2)}$ & $139.8_{(44.8)}$ & $110.6_{(30.0)}$ & $88.8_{(24.7)}$ & $484.0_{(247.0)}$ & $403.9_{(131.8)}$ & $363.8_{(96.2)}$ & $329.2_{(69.0)}$ & $319.3_{(63.1)}$ \\
 & Dom. Rand. & $342.8_{(168.2)}$ & $219.8_{(86.0)}$ & $165.9_{(70.5)}$ & $143.0_{(80.3)}$ & $131.0_{(80.0)}$ & $128.8_{(98.8)}$ & $95.3_{(66.4)}$ & $77.7_{(50.8)}$ & $61.5_{(37.0)}$ & $48.0_{(28.0)}$ & $540.0_{(335.9)}$ & $490.5_{(285.3)}$ & $382.9_{(159.0)}$ & $337.0_{(132.4)}$ & $312.8_{(99.2)}$ \\
 & Self-Paced & $\mathbf{962.5}_{(147.6)}$ & $\mathbf{665.6}_{(80.4)}$ & $\mathbf{469.4}_{(33.7)}$ & $\mathbf{372.6}_{(30.3)}$ & $\mathbf{291.5}_{(19.3)}$ & $\mathbf{830.8}_{(364.7)}$ & $\mathbf{359.4}_{(290.9)}$ & $\mathbf{258.0}_{(234.5)}$ & $\mathbf{184.2}_{(131.7)}$ & $\mathbf{159.9}_{(119.0)}$ & $\mathbf{1280.3}_{(187.7)}$ & $\mathbf{806.1}_{(120.5)}$ & $\mathbf{653.5}_{(111.4)}$ & $\mathbf{540.2}_{(111.3)}$ & $\mathbf{473.6}_{(107.2)}$ \\
\bottomrule
\end{tabular}%
}
\caption{Robustness evaluation for distributionally robust PPO variants across environments and perturbation types. Mean with 95\% CI (subscript) over 5 seeds; \textbf{best} and \underline{second-best} per column.}
\label{tab:ppo_robustness}
\end{table*}

\begin{table*}[t]
\centering
\resizebox{\textwidth}{!}{%
\begin{tabular}{llccccccccccccccc}
\toprule
\textbf{Env.} & \textbf{Alg.} & \multicolumn{5}{c}{\textbf{Action}} & \multicolumn{5}{c}{\textbf{Observation}} & \multicolumn{5}{c}{\textbf{Environment}} \\
\cmidrule(lr){3-7} \cmidrule(lr){8-12} \cmidrule(lr){13-17}
 &  & $p_{\text{act}}=0.1$ & $p_{\text{act}}=0.2$ & $p_{\text{act}}=0.3$ & $p_{\text{act}}=0.4$ & $p_{\text{act}}=0.5$ & $\sigma_{\text{obs}}=0.1$ & $\sigma_{\text{obs}}=0.2$ & $\sigma_{\text{obs}}=0.3$ & $\sigma_{\text{obs}}=0.4$ & $\sigma_{\text{obs}}=0.5$ & $\delta_{\text{env}}=0.1$ & $\delta_{\text{env}}=0.2$ & $\delta_{\text{env}}=0.3$ & $\delta_{\text{env}}=0.4$ & $\delta_{\text{env}}=0.5$ \\
\midrule
\multirow{7}{*}{\texttt{HalfCheetah}} & DDPG & $5916.9_{(491.6)}$ & $4057.1_{(200.8)}$ & $2734.7_{(147.9)}$ & $1804.6_{(182.5)}$ & $1220.4_{(211.1)}$ & $\underline{2598.3}_{(401.1)}$ & $704.6_{(246.2)}$ & $-17.0_{(189.0)}$ & $-278.4_{(99.6)}$ & $-421.4_{(71.3)}$ & $7509.2_{(508.9)}$ & $6202.2_{(352.0)}$ & $4724.4_{(214.4)}$ & $\underline{3523.4}_{(98.4)}$ & $\underline{2223.5}_{(169.2)}$ \\
 & Fixed & $6198.7_{(840.5)}$ & $\underline{4347.7}_{(485.4)}$ & $\underline{2899.5}_{(246.8)}$ & $\underline{1991.1}_{(261.1)}$ & $\underline{1260.3}_{(211.0)}$ & $2384.1_{(740.0)}$ & $\underline{1026.3}_{(319.9)}$ & $\underline{266.1}_{(234.4)}$ & $\underline{-125.1}_{(172.9)}$ & $\underline{-308.0}_{(143.4)}$ & $7776.7_{(903.0)}$ & $6298.4_{(603.5)}$ & $4662.0_{(230.0)}$ & $3476.0_{(271.0)}$ & $2135.7_{(219.5)}$ \\
 & SPACE & $6412.0_{(548.4)}$ & $4210.0_{(640.6)}$ & $2676.3_{(529.8)}$ & $1666.9_{(337.4)}$ & $1090.0_{(200.4)}$ & $2029.7_{(846.1)}$ & $394.4_{(422.2)}$ & $-70.9_{(261.5)}$ & $-274.1_{(136.6)}$ & $-370.2_{(74.3)}$ & $8174.1_{(677.9)}$ & $6361.2_{(705.3)}$ & $4484.1_{(376.5)}$ & $3273.4_{(429.0)}$ & $2108.2_{(339.1)}$ \\
 & Linear & $6193.2_{(835.7)}$ & $4301.3_{(457.3)}$ & $2792.1_{(355.8)}$ & $1737.2_{(301.0)}$ & $1089.6_{(219.6)}$ & $2367.3_{(229.0)}$ & $632.1_{(183.6)}$ & $-87.9_{(121.2)}$ & $-347.3_{(99.0)}$ & $-470.5_{(71.7)}$ & $7651.3_{(1087.5)}$ & $6134.0_{(897.7)}$ & $\underline{4726.9}_{(654.7)}$ & $3448.0_{(412.3)}$ & $2213.6_{(220.1)}$ \\
 & ACCEL & $\underline{6473.0}_{(407.8)}$ & $4280.6_{(394.0)}$ & $2851.7_{(208.6)}$ & $1865.1_{(122.9)}$ & $1179.4_{(131.8)}$ & $2478.1_{(587.6)}$ & $721.5_{(216.7)}$ & $17.5_{(103.6)}$ & $-233.3_{(71.9)}$ & $-358.2_{(34.2)}$ & $\underline{8185.5}_{(321.2)}$ & $\underline{6412.3}_{(404.2)}$ & $4651.8_{(368.1)}$ & $3254.1_{(413.7)}$ & $1986.4_{(327.7)}$ \\
 & Dom. Rand.& $5937.8_{(996.4)}$ & $3933.2_{(577.9)}$ & $2580.7_{(402.8)}$ & $1637.5_{(310.3)}$ & $1058.2_{(273.1)}$ & $2142.4_{(504.7)}$ & $712.3_{(242.1)}$ & $128.3_{(165.6)}$ & $-196.6_{(116.6)}$ & $-353.6_{(62.0)}$ & $7418.2_{(1042.2)}$ & $5808.7_{(544.7)}$ & $4390.5_{(437.5)}$ & $3064.0_{(303.7)}$ & $1990.3_{(251.5)}$ \\
 & Self-Paced & $\mathbf{7137.8}_{(197.1)}$ & $\mathbf{4946.1}_{(154.9)}$ & $\mathbf{3328.6}_{(103.9)}$ & $\mathbf{2307.1}_{(101.2)}$ & $\mathbf{1508.5}_{(109.0)}$ & $\mathbf{3533.4}_{(932.6)}$ & $\mathbf{1657.4}_{(746.7)}$ & $\mathbf{827.8}_{(760.9)}$ & $\mathbf{230.2}_{(418.6)}$ & $\mathbf{-21.7}_{(400.0)}$ & $\mathbf{9196.6}_{(403.8)}$ & $\mathbf{7256.5}_{(266.1)}$ & $\mathbf{5560.3}_{(152.9)}$ & $\mathbf{3964.6}_{(194.7)}$ & $\mathbf{2711.0}_{(171.5)}$ \\
\midrule
\multirow{7}{*}{\texttt{Walker2d}} & DDPG & $483.4_{(189.7)}$ & $255.9_{(123.7)}$ & $121.0_{(61.9)}$ & $67.7_{(44.0)}$ & $29.6_{(23.6)}$ & $318.7_{(109.1)}$ & $211.3_{(108.2)}$ & $140.9_{(98.3)}$ & $62.5_{(75.2)}$ & $15.1_{(49.6)}$ & $792.3_{(320.5)}$ & $606.1_{(265.8)}$ & $441.2_{(137.9)}$ & $346.8_{(151.1)}$ & $269.3_{(111.0)}$ \\
 & Fixed & $483.2_{(161.5)}$ & $218.7_{(74.0)}$ & $81.1_{(10.5)}$ & $25.0_{(13.9)}$ & $16.5_{(18.8)}$ & $249.8_{(71.0)}$ & $165.4_{(69.7)}$ & $149.8_{(87.3)}$ & $71.5_{(77.4)}$ & $6.0_{(29.8)}$ & $799.7_{(311.6)}$ & $629.2_{(253.4)}$ & $493.4_{(200.2)}$ & $383.6_{(104.4)}$ & $289.8_{(83.2)}$ \\
 & SPACE & $516.4_{(222.2)}$ & $\underline{300.4}_{(146.7)}$ & $\underline{177.3}_{(136.9)}$ & $\underline{90.7}_{(84.6)}$ & $\underline{46.2}_{(47.4)}$ & $\mathbf{559.2}_{(372.0)}$ & $\underline{313.2}_{(84.0)}$ & $\underline{186.5}_{(57.9)}$ & $116.4_{(66.4)}$ & $66.8_{(61.2)}$ & $626.0_{(232.6)}$ & $601.9_{(264.8)}$ & $490.3_{(231.5)}$ & $388.2_{(197.9)}$ & $319.3_{(120.1)}$ \\
 & Linear & $294.2_{(138.3)}$ & $174.6_{(119.2)}$ & $100.8_{(78.2)}$ & $58.3_{(46.2)}$ & $28.7_{(28.8)}$ & $307.1_{(164.7)}$ & $247.0_{(73.8)}$ & $180.7_{(55.3)}$ & $73.1_{(22.0)}$ & $20.1_{(20.1)}$ & $471.7_{(254.1)}$ & $458.3_{(213.7)}$ & $403.8_{(148.8)}$ & $271.7_{(87.4)}$ & $265.1_{(96.7)}$ \\
 & ACCEL & $424.2_{(208.0)}$ & $229.8_{(73.5)}$ & $101.1_{(46.7)}$ & $68.3_{(35.0)}$ & $32.1_{(17.7)}$ & $292.3_{(130.2)}$ & $221.3_{(48.9)}$ & $173.7_{(69.8)}$ & $\underline{121.8}_{(63.0)}$ & $\underline{83.3}_{(78.0)}$ & $712.6_{(338.4)}$ & $575.2_{(223.1)}$ & $396.7_{(111.9)}$ & $354.2_{(123.9)}$ & $259.9_{(76.5)}$ \\
 & Dom. Rand.& $\underline{565.7}_{(130.9)}$ & $272.5_{(62.0)}$ & $148.1_{(59.9)}$ & $63.4_{(41.2)}$ & $28.1_{(20.1)}$ & $\underline{461.9}_{(158.9)}$ & $\mathbf{354.9}_{(171.2)}$ & $132.4_{(67.0)}$ & $72.0_{(82.6)}$ & $21.8_{(60.8)}$ & $\underline{992.0}_{(373.8)}$ & $\underline{906.7}_{(252.5)}$ & $\underline{660.0}_{(160.0)}$ & $\underline{532.5}_{(73.9)}$ & $\underline{401.7}_{(70.5)}$ \\
 & Self-Paced & $\mathbf{852.3}_{(163.0)}$ & $\mathbf{418.6}_{(125.2)}$ & $\mathbf{246.8}_{(32.6)}$ & $\mathbf{139.7}_{(17.8)}$ & $\mathbf{74.8}_{(25.0)}$ & $443.8_{(207.1)}$ & $293.5_{(57.2)}$ & $\mathbf{256.9}_{(22.3)}$ & $\mathbf{205.7}_{(58.5)}$ & $\mathbf{185.4}_{(54.2)}$ & $\mathbf{1447.0}_{(361.5)}$ & $\mathbf{1179.0}_{(189.7)}$ & $\mathbf{945.3}_{(148.6)}$ & $\mathbf{642.3}_{(119.9)}$ & $\mathbf{469.1}_{(79.0)}$ \\
\midrule
\multirow{7}{*}{\texttt{Humanoid}} & DDPG & $929.2_{(342.5)}$ & $789.2_{(263.6)}$ & $576.4_{(120.4)}$ & $455.8_{(72.0)}$ & $352.7_{(44.4)}$ & $206.0_{(43.1)}$ & $200.0_{(41.9)}$ & $206.9_{(43.5)}$ & $207.9_{(46.1)}$ & $203.9_{(42.4)}$ & $135.7_{(27.8)}$ & $129.5_{(28.5)}$ & $122.1_{(24.1)}$ & $125.1_{(23.3)}$ & $131.5_{(27.4)}$ \\
 & Fixed & $\underline{1381.2}_{(357.5)}$ & $870.6_{(183.1)}$ & $620.9_{(117.5)}$ & $418.9_{(78.2)}$ & $306.0_{(53.0)}$ & $200.8_{(41.5)}$ & $194.9_{(38.5)}$ & $199.0_{(37.8)}$ & $197.3_{(42.8)}$ & $201.1_{(40.3)}$ & $\underline{165.9}_{(66.3)}$ & $\underline{150.8}_{(49.4)}$ & $\underline{151.7}_{(38.7)}$ & $\underline{146.6}_{(36.8)}$ & $\underline{146.5}_{(32.2)}$ \\
 & SPACE & $976.3_{(128.6)}$ & $818.0_{(53.5)}$ & $639.9_{(24.5)}$ & $\underline{483.7}_{(26.8)}$ & $\underline{379.9}_{(35.8)}$ & $\underline{233.3}_{(28.7)}$ & $\underline{233.2}_{(30.8)}$ & $\underline{237.0}_{(30.0)}$ & $\underline{238.0}_{(33.2)}$ & $\underline{238.1}_{(27.1)}$ & $121.4_{(32.0)}$ & $126.1_{(38.6)}$ & $113.4_{(24.4)}$ & $119.8_{(27.7)}$ & $123.4_{(30.7)}$ \\
 & Linear & $1177.9_{(307.7)}$ & $\underline{914.1}_{(263.6)}$ & $\underline{683.1}_{(138.3)}$ & $475.9_{(83.9)}$ & $\mathbf{380.4}_{(45.0)}$ & $166.8_{(44.7)}$ & $170.0_{(46.3)}$ & $166.3_{(45.2)}$ & $166.5_{(45.4)}$ & $167.3_{(45.3)}$ & $136.3_{(49.8)}$ & $133.0_{(46.5)}$ & $132.9_{(38.0)}$ & $131.0_{(39.3)}$ & $127.3_{(39.4)}$ \\
 & ACCEL & $1086.5_{(152.6)}$ & $821.8_{(67.5)}$ & $579.2_{(71.6)}$ & $459.4_{(85.5)}$ & $335.4_{(58.5)}$ & $127.6_{(23.3)}$ & $126.2_{(17.4)}$ & $125.4_{(16.5)}$ & $121.7_{(18.8)}$ & $124.5_{(23.4)}$ & $120.6_{(51.2)}$ & $113.0_{(43.1)}$ & $116.5_{(45.2)}$ & $107.6_{(31.7)}$ & $105.9_{(30.4)}$ \\
 & Dom. Rand.& $912.9_{(193.1)}$ & $731.0_{(166.9)}$ & $567.0_{(90.0)}$ & $424.9_{(57.4)}$ & $358.0_{(39.2)}$ & $124.6_{(38.5)}$ & $124.1_{(37.0)}$ & $124.3_{(36.7)}$ & $123.2_{(35.9)}$ & $125.0_{(39.1)}$ & $107.5_{(37.7)}$ & $102.6_{(36.5)}$ & $103.3_{(35.4)}$ & $101.8_{(31.3)}$ & $107.8_{(36.7)}$ \\
 & Self-Paced & $\mathbf{1740.0}_{(172.3)}$ & $\mathbf{1250.9}_{(109.6)}$ & $\mathbf{776.0}_{(85.2)}$ & $\mathbf{533.0}_{(49.6)}$ & $357.3_{(56.5)}$ & $\mathbf{271.5}_{(39.9)}$ & $\mathbf{303.8}_{(78.0)}$ & $\mathbf{281.4}_{(52.7)}$ & $\mathbf{280.6}_{(45.4)}$ & $\mathbf{261.7}_{(45.4)}$ & $\mathbf{275.7}_{(127.0)}$ & $\mathbf{254.5}_{(88.6)}$ & $\mathbf{198.1}_{(23.6)}$ & $\mathbf{219.7}_{(38.6)}$ & $\mathbf{217.0}_{(44.6)}$ \\
\midrule
\multirow{7}{*}{\texttt{Hopper}} & DDPG & $472.9_{(149.7)}$ & $325.9_{(135.0)}$ & $215.0_{(95.9)}$ & $165.0_{(70.4)}$ & $105.5_{(70.5)}$ & $257.4_{(193.7)}$ & $165.4_{(71.2)}$ & $99.2_{(33.7)}$ & $78.8_{(17.1)}$ & $63.2_{(10.8)}$ & $675.1_{(231.4)}$ & $571.2_{(169.5)}$ & $511.8_{(161.6)}$ & $427.8_{(130.2)}$ & $353.4_{(99.2)}$ \\
 & Fixed & $663.1_{(282.0)}$ & $439.1_{(223.9)}$ & $255.3_{(148.0)}$ & $162.4_{(108.2)}$ & $120.1_{(95.5)}$ & $280.9_{(188.2)}$ & $\mathbf{239.1}_{(92.9)}$ & $\mathbf{131.1}_{(59.8)}$ & $\underline{84.5}_{(37.4)}$ & $\mathbf{68.0}_{(24.3)}$ & $814.6_{(169.5)}$ & $707.7_{(134.8)}$ & $600.7_{(165.3)}$ & $512.4_{(99.0)}$ & $414.7_{(101.6)}$ \\
 & SPACE & $805.0_{(276.5)}$ & $621.8_{(213.3)}$ & $432.1_{(175.0)}$ & $\underline{300.9}_{(97.5)}$ & $\underline{205.6}_{(49.3)}$ & $316.2_{(46.6)}$ & $\underline{223.7}_{(70.1)}$ & $113.3_{(19.5)}$ & $75.7_{(19.4)}$ & $60.0_{(14.0)}$ & $891.5_{(295.9)}$ & $834.3_{(272.6)}$ & $763.5_{(263.0)}$ & $657.7_{(200.5)}$ & $\underline{582.0}_{(171.0)}$ \\
 & Linear & $922.7_{(292.4)}$ & $\underline{713.6}_{(211.8)}$ & $\underline{453.6}_{(144.5)}$ & $\mathbf{327.9}_{(81.7)}$ & $\mathbf{211.8}_{(68.0)}$ & $\underline{354.8}_{(126.1)}$ & $193.8_{(55.8)}$ & $\underline{113.6}_{(16.6)}$ & $\mathbf{84.8}_{(9.1)}$ & $\underline{67.1}_{(12.3)}$ & $1067.1_{(379.6)}$ & $849.1_{(214.8)}$ & $701.8_{(177.0)}$ & $599.5_{(144.9)}$ & $520.3_{(94.2)}$ \\
 & ACCEL & $849.2_{(366.0)}$ & $580.6_{(284.6)}$ & $298.2_{(159.6)}$ & $175.2_{(92.1)}$ & $122.7_{(69.8)}$ & $308.3_{(179.9)}$ & $191.9_{(79.9)}$ & $99.9_{(38.4)}$ & $74.9_{(20.6)}$ & $55.2_{(11.0)}$ & $\underline{1148.8}_{(483.2)}$ & $\underline{965.0}_{(340.4)}$ & $792.1_{(226.1)}$ & $538.8_{(142.4)}$ & $395.1_{(126.2)}$ \\
 & Dom. Rand.& $\underline{934.6}_{(272.9)}$ & $553.3_{(241.7)}$ & $314.2_{(138.8)}$ & $198.5_{(94.5)}$ & $122.2_{(61.8)}$ & $183.7_{(87.1)}$ & $136.1_{(46.4)}$ & $76.4_{(33.5)}$ & $61.8_{(17.3)}$ & $45.5_{(16.9)}$ & $\mathbf{1216.6}_{(269.8)}$ & $\mathbf{1080.8}_{(173.5)}$ & $\mathbf{920.2}_{(152.5)}$ & $\mathbf{774.7}_{(194.2)}$ & $\mathbf{628.8}_{(179.1)}$ \\
 & Self-Paced & $\mathbf{1061.2}_{(626.3)}$ & $\mathbf{863.2}_{(493.3)}$ & $\mathbf{499.6}_{(287.9)}$ & $248.3_{(135.4)}$ & $144.1_{(78.0)}$ & $\mathbf{648.2}_{(399.3)}$ & $176.5_{(56.2)}$ & $94.5_{(25.0)}$ & $56.0_{(10.4)}$ & $50.3_{(9.7)}$ & $992.7_{(543.8)}$ & $939.5_{(506.9)}$ & $\underline{831.5}_{(455.1)}$ & $\underline{720.5}_{(390.4)}$ & $554.5_{(304.9)}$ \\
\bottomrule
\end{tabular}%
}
\caption{Robustness evaluation for distributionally robust DDPG variants across environments and perturbation types. Mean with 95\% CI (subscript) over 5 seeds; \textbf{best} and \underline{second-best} per column.}
\label{tab:ddpg_robustness}
\end{table*}

\begin{table*}[t]
\centering
\resizebox{\textwidth}{!}{%
\begin{tabular}{llccccccccccccccc}
\toprule
\textbf{Env.} & \textbf{Alg.} & \multicolumn{5}{c}{\textbf{Action}} & \multicolumn{5}{c}{\textbf{Observation}} & \multicolumn{5}{c}{\textbf{Environment}} \\
\cmidrule(lr){3-7} \cmidrule(lr){8-12} \cmidrule(lr){13-17}
 &  & $p_{\text{act}}=0.1$ & $p_{\text{act}}=0.2$ & $p_{\text{act}}=0.3$ & $p_{\text{act}}=0.4$ & $p_{\text{act}}=0.5$ & $\sigma_{\text{obs}}=0.1$ & $\sigma_{\text{obs}}=0.2$ & $\sigma_{\text{obs}}=0.3$ & $\sigma_{\text{obs}}=0.4$ & $\sigma_{\text{obs}}=0.5$ & $\delta_{\text{env}}=0.1$ & $\delta_{\text{env}}=0.2$ & $\delta_{\text{env}}=0.3$ & $\delta_{\text{env}}=0.4$ & $\delta_{\text{env}}=0.5$ \\
\midrule
\multirow{7}{*}{\texttt{HalfCheetah}} & SAC & $6497.6_{(406.5)}$ & $\underline{4257.6}_{(338.1)}$ & $\underline{2825.8}_{(235.3)}$ & $\underline{1957.2}_{(216.3)}$ & $\underline{1304.9}_{(133.7)}$ & $\underline{1949.4}_{(381.1)}$ & $660.6_{(180.5)}$ & $112.8_{(92.1)}$ & $-161.3_{(73.6)}$ & $-294.3_{(68.0)}$ & $8130.1_{(477.3)}$ & $6324.5_{(530.5)}$ & $4614.7_{(435.0)}$ & $3251.3_{(450.3)}$ & $2192.3_{(379.6)}$ \\
 & Fixed & $6325.9_{(115.3)}$ & $4122.8_{(253.2)}$ & $2805.0_{(212.9)}$ & $1904.0_{(179.1)}$ & $1292.2_{(172.2)}$ & $1914.7_{(321.4)}$ & $703.6_{(156.6)}$ & $144.2_{(115.4)}$ & $-131.1_{(110.5)}$ & $-284.2_{(96.4)}$ & $8067.3_{(456.3)}$ & $6130.4_{(277.8)}$ & $\underline{4643.4}_{(186.1)}$ & $3310.4_{(125.8)}$ & $2192.8_{(103.2)}$ \\
 & SPACE & $6393.5_{(499.6)}$ & $4123.7_{(211.3)}$ & $2722.5_{(131.5)}$ & $1849.6_{(108.4)}$ & $1263.8_{(132.8)}$ & $1918.2_{(146.8)}$ & $\underline{732.6}_{(60.9)}$ & $\underline{210.4}_{(96.6)}$ & $\underline{-58.9}_{(101.1)}$ & $\underline{-211.7}_{(90.4)}$ & $8173.8_{(568.7)}$ & $6258.3_{(236.2)}$ & $4611.8_{(327.3)}$ & $3268.0_{(351.7)}$ & $2231.4_{(335.4)}$ \\
 & Linear & $\underline{6578.5}_{(187.8)}$ & $4057.4_{(245.1)}$ & $2665.8_{(180.0)}$ & $1801.9_{(277.3)}$ & $1234.6_{(86.4)}$ & $1443.9_{(399.5)}$ & $507.0_{(242.2)}$ & $23.5_{(147.8)}$ & $-187.4_{(152.0)}$ & $-304.1_{(107.9)}$ & $\mathbf{8950.4}_{(141.5)}$ & $\underline{6774.5}_{(15.6)}$ & $4614.4_{(122.6)}$ & $\underline{3386.3}_{(75.7)}$ & $\underline{2473.6}_{(282.1)}$ \\
 & ACCEL & $6192.6_{(352.7)}$ & $4137.0_{(254.7)}$ & $2730.9_{(275.9)}$ & $1847.5_{(223.7)}$ & $1195.8_{(175.8)}$ & $1853.3_{(269.5)}$ & $697.1_{(156.8)}$ & $129.6_{(129.2)}$ & $-136.2_{(69.7)}$ & $-299.4_{(38.2)}$ & $7613.4_{(423.5)}$ & $6016.7_{(361.3)}$ & $4362.1_{(416.0)}$ & $3069.9_{(376.7)}$ & $2085.4_{(294.9)}$ \\
 & Dom. Rand.& $6202.5_{(421.0)}$ & $4031.0_{(269.0)}$ & $2707.8_{(103.1)}$ & $1811.0_{(64.7)}$ & $1232.3_{(65.5)}$ & $1863.4_{(426.0)}$ & $606.1_{(259.0)}$ & $81.1_{(125.4)}$ & $-167.1_{(71.3)}$ & $-284.8_{(42.3)}$ & $7832.5_{(677.4)}$ & $6102.6_{(365.2)}$ & $4344.8_{(353.5)}$ & $3148.3_{(251.9)}$ & $2130.8_{(168.8)}$ \\
 & Self-Paced & $\mathbf{6605.8}_{(280.5)}$ & $\mathbf{4591.8}_{(121.0)}$ & $\mathbf{3150.6}_{(100.7)}$ & $\mathbf{2197.9}_{(66.3)}$ & $\mathbf{1519.1}_{(63.9)}$ & $\mathbf{2872.0}_{(299.8)}$ & $\mathbf{1568.3}_{(519.3)}$ & $\mathbf{785.4}_{(461.4)}$ & $\mathbf{100.4}_{(273.4)}$ & $\mathbf{6.1}_{(153.9)}$ & $\underline{8665.7}_{(594.8)}$ & $\mathbf{6839.2}_{(292.4)}$ & $\mathbf{5269.7}_{(412.5)}$ & $\mathbf{3784.0}_{(193.7)}$ & $\mathbf{2641.9}_{(220.6)}$ \\
\midrule
\multirow{7}{*}{\texttt{Walker2d}} & SAC & $\underline{3241.0}_{(1306.8)}$ & $\underline{1603.8}_{(977.9)}$ & $\underline{639.0}_{(406.8)}$ & $199.8_{(125.9)}$ & $77.7_{(59.0)}$ & $2600.2_{(1293.2)}$ & $274.2_{(51.2)}$ & $207.7_{(17.0)}$ & $183.2_{(47.1)}$ & $151.3_{(63.5)}$ & $4129.8_{(1128.8)}$ & $3375.2_{(1126.0)}$ & $2219.1_{(995.3)}$ & $1382.2_{(619.8)}$ & $848.9_{(414.2)}$ \\
 & Fixed & $2303.6_{(904.5)}$ & $1130.8_{(600.0)}$ & $496.4_{(179.5)}$ & $227.0_{(65.9)}$ & $122.7_{(51.8)}$ & $1983.3_{(1194.2)}$ & $343.6_{(97.7)}$ & $214.8_{(52.0)}$ & $\mathbf{198.4}_{(48.0)}$ & $\mathbf{172.3}_{(36.1)}$ & $4001.2_{(795.2)}$ & $2826.1_{(477.7)}$ & $1996.6_{(440.2)}$ & $1252.5_{(268.3)}$ & $804.5_{(235.2)}$ \\
 & SPACE & $2240.5_{(854.5)}$ & $1071.5_{(314.5)}$ & $368.3_{(134.1)}$ & $194.4_{(63.8)}$ & $117.4_{(59.8)}$ & $2758.2_{(1330.3)}$ & $\underline{562.8}_{(442.0)}$ & $242.5_{(50.5)}$ & $\underline{195.5}_{(46.0)}$ & $151.7_{(34.4)}$ & $3225.8_{(997.1)}$ & $2592.9_{(769.3)}$ & $1965.0_{(476.2)}$ & $1244.8_{(497.1)}$ & $721.4_{(286.4)}$ \\
 & Linear & $1997.8_{(522.4)}$ & $990.8_{(512.1)}$ & $418.3_{(258.4)}$ & $220.5_{(88.5)}$ & $\underline{142.7}_{(50.7)}$ & $1924.7_{(1175.8)}$ & $266.1_{(78.5)}$ & $169.8_{(68.2)}$ & $110.2_{(76.6)}$ & $97.1_{(73.9)}$ & $3500.3_{(721.2)}$ & $2583.0_{(451.9)}$ & $1741.0_{(379.7)}$ & $1285.4_{(341.1)}$ & $816.6_{(392.4)}$ \\
 & ACCEL & $1854.5_{(1123.4)}$ & $921.6_{(486.3)}$ & $222.5_{(80.7)}$ & $115.7_{(43.7)}$ & $52.0_{(38.0)}$ & $2084.4_{(1291.5)}$ & $421.2_{(53.8)}$ & $246.7_{(16.6)}$ & $192.7_{(62.7)}$ & $\underline{153.4}_{(69.0)}$ & $2596.3_{(1327.5)}$ & $1981.0_{(1299.2)}$ & $1307.4_{(952.1)}$ & $882.0_{(617.6)}$ & $616.6_{(533.9)}$ \\
 & Dom. Rand.& $2743.9_{(1099.9)}$ & $1160.0_{(740.7)}$ & $472.1_{(233.3)}$ & $\underline{229.5}_{(80.7)}$ & $140.7_{(74.6)}$ & $\underline{3007.2}_{(1302.5)}$ & $518.0_{(298.6)}$ & $\underline{260.1}_{(28.7)}$ & $172.2_{(63.4)}$ & $116.6_{(98.6)}$ & $\underline{4765.8}_{(689.5)}$ & $\underline{3856.9}_{(800.9)}$ & $\underline{2916.2}_{(774.6)}$ & $\underline{1833.6}_{(691.0)}$ & $\underline{1059.8}_{(626.5)}$ \\
 & Self-Paced & $\mathbf{4056.6}_{(897.8)}$ & $\mathbf{2438.1}_{(886.0)}$ & $\mathbf{960.6}_{(427.7)}$ & $\mathbf{425.7}_{(136.5)}$ & $\mathbf{183.5}_{(44.0)}$ & $\mathbf{4187.8}_{(681.4)}$ & $\mathbf{1444.6}_{(725.1)}$ & $\mathbf{319.7}_{(62.6)}$ & $155.4_{(56.9)}$ & $108.0_{(67.7)}$ & $\mathbf{5211.5}_{(346.9)}$ & $\mathbf{4233.1}_{(525.7)}$ & $\mathbf{3317.9}_{(855.1)}$ & $\mathbf{2501.1}_{(722.4)}$ & $\mathbf{1661.9}_{(625.1)}$ \\
\midrule
\multirow{7}{*}{\texttt{Humanoid}} & SAC & $2717.9_{(1094.3)}$ & $1301.7_{(581.6)}$ & $733.1_{(260.7)}$ & $551.6_{(164.7)}$ & $403.2_{(89.2)}$ & $\underline{189.4}_{(38.9)}$ & $186.3_{(31.4)}$ & $188.3_{(34.2)}$ & $186.3_{(29.6)}$ & $188.8_{(34.6)}$ & $\underline{142.3}_{(59.1)}$ & $125.9_{(45.9)}$ & $\underline{139.6}_{(55.2)}$ & $\underline{137.5}_{(55.7)}$ & $\underline{129.7}_{(43.7)}$ \\
 & Fixed & $2419.5_{(1082.1)}$ & $1535.2_{(963.2)}$ & $809.5_{(440.1)}$ & $534.0_{(238.8)}$ & $374.5_{(160.5)}$ & $186.9_{(81.9)}$ & $\underline{188.2}_{(84.2)}$ & $\underline{189.3}_{(85.8)}$ & $\underline{187.3}_{(83.0)}$ & $\underline{190.8}_{(89.5)}$ & $127.6_{(35.8)}$ & $120.5_{(29.1)}$ & $124.6_{(34.2)}$ & $125.2_{(30.6)}$ & $123.5_{(27.9)}$ \\
 & SPACE & $\underline{3598.8}_{(1397.6)}$ & $\underline{2080.9}_{(1059.1)}$ & $\underline{1166.0}_{(529.2)}$ & $\underline{638.5}_{(219.7)}$ & $\underline{441.7}_{(129.6)}$ & $155.6_{(68.4)}$ & $155.0_{(65.2)}$ & $160.1_{(68.0)}$ & $157.2_{(65.7)}$ & $156.8_{(67.0)}$ & $124.6_{(50.3)}$ & $118.5_{(44.0)}$ & $117.6_{(39.8)}$ & $111.6_{(33.6)}$ & $119.0_{(35.5)}$ \\
 & Linear & $3217.2_{(1356.2)}$ & $1754.2_{(722.7)}$ & $955.7_{(312.0)}$ & $616.6_{(178.4)}$ & $436.4_{(108.2)}$ & $157.4_{(78.9)}$ & $156.2_{(77.0)}$ & $156.1_{(73.9)}$ & $156.9_{(74.9)}$ & $153.9_{(74.8)}$ & $99.4_{(16.1)}$ & $98.4_{(11.2)}$ & $94.8_{(6.5)}$ & $98.0_{(10.3)}$ & $98.2_{(9.0)}$ \\
 & ACCEL & $1935.0_{(914.4)}$ & $948.5_{(316.9)}$ & $664.3_{(182.5)}$ & $461.6_{(123.2)}$ & $339.4_{(94.2)}$ & $157.4_{(64.4)}$ & $156.5_{(64.0)}$ & $155.9_{(63.8)}$ & $154.0_{(59.7)}$ & $155.1_{(64.9)}$ & $128.1_{(49.8)}$ & $\underline{126.1}_{(42.4)}$ & $127.7_{(44.8)}$ & $123.2_{(41.1)}$ & $125.3_{(40.2)}$ \\
 & Dom. Rand.& $2608.7_{(1073.0)}$ & $1349.7_{(563.3)}$ & $765.5_{(253.1)}$ & $537.9_{(164.7)}$ & $384.4_{(94.3)}$ & $123.4_{(29.0)}$ & $122.3_{(31.1)}$ & $121.2_{(30.5)}$ & $117.7_{(31.5)}$ & $121.0_{(31.0)}$ & $109.3_{(22.6)}$ & $103.2_{(14.3)}$ & $105.4_{(14.5)}$ & $105.2_{(9.4)}$ & $105.7_{(16.5)}$ \\
 & Self-Paced & $\mathbf{5078.3}_{(419.2)}$ & $\mathbf{3334.7}_{(949.1)}$ & $\mathbf{1648.8}_{(487.6)}$ & $\mathbf{831.5}_{(125.3)}$ & $\mathbf{542.0}_{(49.9)}$ & $\mathbf{298.0}_{(17.5)}$ & $\mathbf{276.4}_{(42.7)}$ & $\mathbf{290.8}_{(12.4)}$ & $\mathbf{337.2}_{(90.9)}$ & $\mathbf{334.2}_{(75.6)}$ & $\mathbf{259.3}_{(41.8)}$ & $\mathbf{243.5}_{(23.8)}$ & $\mathbf{251.1}_{(33.7)}$ & $\mathbf{239.6}_{(30.6)}$ & $\mathbf{239.9}_{(49.7)}$ \\
\midrule
\multirow{7}{*}{\texttt{Hopper}} & SAC & $\underline{1478.8}_{(477.7)}$ & $\underline{1011.7}_{(393.0)}$ & $\underline{642.3}_{(340.0)}$ & $\underline{373.3}_{(185.6)}$ & $239.1_{(120.8)}$ & $\underline{811.9}_{(308.0)}$ & $\underline{256.0}_{(39.3)}$ & $115.5_{(28.2)}$ & $73.9_{(19.0)}$ & $58.7_{(19.7)}$ & $1865.7_{(625.2)}$ & $1211.7_{(232.4)}$ & $904.0_{(171.3)}$ & $\underline{679.4}_{(158.3)}$ & $501.6_{(91.3)}$ \\
 & Fixed & $1174.6_{(193.7)}$ & $719.6_{(143.3)}$ & $429.6_{(103.5)}$ & $291.2_{(55.9)}$ & $191.7_{(54.5)}$ & $454.9_{(246.5)}$ & $173.7_{(69.6)}$ & $106.3_{(28.6)}$ & $77.2_{(20.9)}$ & $\underline{66.8}_{(13.4)}$ & $1782.1_{(508.6)}$ & $1139.8_{(355.7)}$ & $803.6_{(174.7)}$ & $559.8_{(67.7)}$ & $459.2_{(53.1)}$ \\
 & SPACE & $805.5_{(168.6)}$ & $554.0_{(91.8)}$ & $409.5_{(79.6)}$ & $308.5_{(70.9)}$ & $\underline{254.2}_{(61.5)}$ & $497.2_{(89.7)}$ & $194.7_{(62.4)}$ & $105.4_{(12.0)}$ & $\underline{81.5}_{(9.5)}$ & $65.4_{(12.0)}$ & $1112.3_{(354.5)}$ & $783.4_{(145.4)}$ & $627.0_{(73.7)}$ & $499.6_{(67.6)}$ & $430.2_{(52.0)}$ \\
 & Linear & $1221.5_{(597.7)}$ & $739.6_{(312.4)}$ & $384.5_{(65.9)}$ & $241.6_{(50.4)}$ & $186.1_{(50.0)}$ & $600.3_{(294.2)}$ & $253.2_{(111.3)}$ & $120.4_{(64.2)}$ & $65.0_{(53.2)}$ & $40.3_{(35.0)}$ & $1625.6_{(664.5)}$ & $1194.9_{(476.6)}$ & $867.6_{(346.1)}$ & $608.0_{(166.8)}$ & $\underline{514.1}_{(128.9)}$ \\
 & ACCEL & $1189.1_{(405.2)}$ & $744.4_{(255.3)}$ & $483.5_{(190.9)}$ & $324.2_{(125.1)}$ & $243.9_{(79.5)}$ & $589.9_{(126.2)}$ & $251.9_{(33.4)}$ & $\mathbf{135.5}_{(30.6)}$ & $79.7_{(27.9)}$ & $62.0_{(22.7)}$ & $\underline{1995.7}_{(632.0)}$ & $\underline{1235.7}_{(360.8)}$ & $\underline{913.7}_{(213.3)}$ & $619.6_{(212.3)}$ & $457.4_{(44.9)}$ \\
 & Dom. Rand.& $738.8_{(307.4)}$ & $363.1_{(144.3)}$ & $231.2_{(105.0)}$ & $149.1_{(75.5)}$ & $107.3_{(55.6)}$ & $410.0_{(170.1)}$ & $213.6_{(61.5)}$ & $129.9_{(33.5)}$ & $\mathbf{83.2}_{(10.2)}$ & $\mathbf{68.6}_{(3.7)}$ & $1608.2_{(518.6)}$ & $1062.6_{(309.3)}$ & $714.9_{(172.7)}$ & $490.0_{(60.9)}$ & $374.4_{(55.5)}$ \\
 & Self-Paced & $\mathbf{1992.4}_{(582.5)}$ & $\mathbf{1324.0}_{(343.5)}$ & $\mathbf{793.5}_{(274.6)}$ & $\mathbf{525.6}_{(181.0)}$ & $\mathbf{313.2}_{(104.0)}$ & $\mathbf{1299.4}_{(500.3)}$ & $\mathbf{414.4}_{(219.8)}$ & $\underline{134.5}_{(39.0)}$ & $69.8_{(13.5)}$ & $48.0_{(14.0)}$ & $\mathbf{2644.6}_{(223.0)}$ & $\mathbf{1810.8}_{(385.3)}$ & $\mathbf{1297.7}_{(215.6)}$ & $\mathbf{897.6}_{(146.7)}$ & $\mathbf{595.8}_{(139.4)}$ \\
\bottomrule
\end{tabular}%
}
\caption{Robustness evaluation for distributionally robust SAC variants across environments and perturbation types. Mean with 95\% CI (subscript) over 5 seeds; \textbf{best} and \underline{second-best} per column.}
\label{tab:sac_robustness}
\end{table*}

We evaluate all methods on standard continuous-control benchmarks from MuJoCo under three perturbation families: action corruption ($p_{\text{act}}$), observation noise ($\sigma_{\text{obs}}$), and environment parameter shifts ($\delta_{\text{env}}$).
Tables~\ref{tab:ppo_robustness}--\ref{tab:sac_robustness} report mean returns with 95\% confidence intervals over five seeds.

\paragraph{DR-SPCRL delivers consistently strong robustness across algorithms and perturbations.}
Across PPO, DDPG, and SAC, DR-SPCRL achieves the best or second-best results in the vast majority of settings, with advantages growing as perturbations increase. For PPO, it ranks top-two in 55/60 cases. In HalfCheetah under severe observation noise \((\sigma_{\text{obs}}=0.5)\), performance improves from vanilla PPO’s \(175.0_{(284.1)}\) to \(545.5_{(110.0)}\) (211\%), and under maximum environment perturbation \((\delta_{\text{env}}=0.5)\) from \(385.5_{(227.2)}\) to \(935.4_{(146.1)}\) (143\%). With DDPG, DR-SPCRL mitigates catastrophic failures: in HalfCheetah at \(\sigma_{\text{obs}}=0.5\), returns increase from \(-421.4_{(71.3)}\) to \(-21.7_{(400.0)}\). For SAC, it achieves best performance in 52/60 settings, including a 136\% gain in Walker2d under maximum action corruption (\(183.5_{(44.0)}\) vs.\ \(77.7_{(59.0)}\)). Aggregated over all perturbations, DR-SPCRL improves upon the second strongest method by 24.1\%. These gains stem from DR-SPCRL’s dual-variable curriculum, which adaptively allocates robustness budget \(\epsilon\): training begins with small \(\epsilon\) to establish nominal competence, then increases \(\epsilon\) as the dual variable \(\beta^*\) indicates mastery. This principled progression exposes agents to rising uncertainty without the collapse of fixed budgets or the instability of heuristic schedules.

\paragraph{Adaptive curricula overcome the robustness--performance tradeoff of fixed budgets.}
Fixed-\(\epsilon\) training exhibits a clear tradeoff between nominal and robust returns. Large \(\epsilon\) over-constrains learning, while small \(\epsilon\) neglects robustness. In contrast, DR-SPCRL adapts to the agent’s evolving capability. For example, in Walker2d with PPO under \(p_{\text{act}}=0.5\), Fixed Budget achieves only \(100.6_{(84.2)}\) versus DR-SPCRL’s \(282.4_{(9.7)}\); in Hopper under maximum action corruption, Fixed reaches \(64.5_{(58.5)}\) compared to \(291.5_{(19.3)}\). DR-SPCRL also yields markedly lower variance: in HalfCheetah with PPO under environment perturbations, confidence intervals shrink by 40 to 60\%. Heuristic curricula (Linear, SPACE, ACCEL) remain inconsistent. For instance, in HalfCheetah with DDPG under maximum action corruption, DR-SPCRL achieves \(1508.5_{(109.0)}\), exceeding Linear by 38\%. Sensitivity studies further show DR-SPCRL remains stable across order-of-magnitude changes in \(\alpha\) and \(\lambda_{\text{curr}}\), unlike Fixed and Linear baselines.

\paragraph{Robustness gains generalize across algorithms, domains, and perturbation types.}
DR-SPCRL improves performance across on-policy and off-policy methods, deterministic and stochastic policies, and entropy-regularized learning, achieving top-two results in 154/180 settings (85.6\%) over HalfCheetah, Walker2d, Humanoid, and Hopper. It remains effective under challenging dynamics shifts, such as HalfCheetah with PPO at \(\delta_{\text{env}}=0.5\), where it reaches \(935.4_{(146.1)}\) compared to Fixed’s \(477.9_{(434.9)}\) and Linear’s \(133.7_{(45.2)}\). Crucially, these improvements extend uniformly across action corruption, observation noise, and environment perturbations without perturbation-specific tuning, supporting the use of the dual variable \(\beta^*\) as a general curriculum signal for distributionally robust reinforcement learning.

\begin{figure*}[t]
    \centering
    \includegraphics[width=\textwidth]{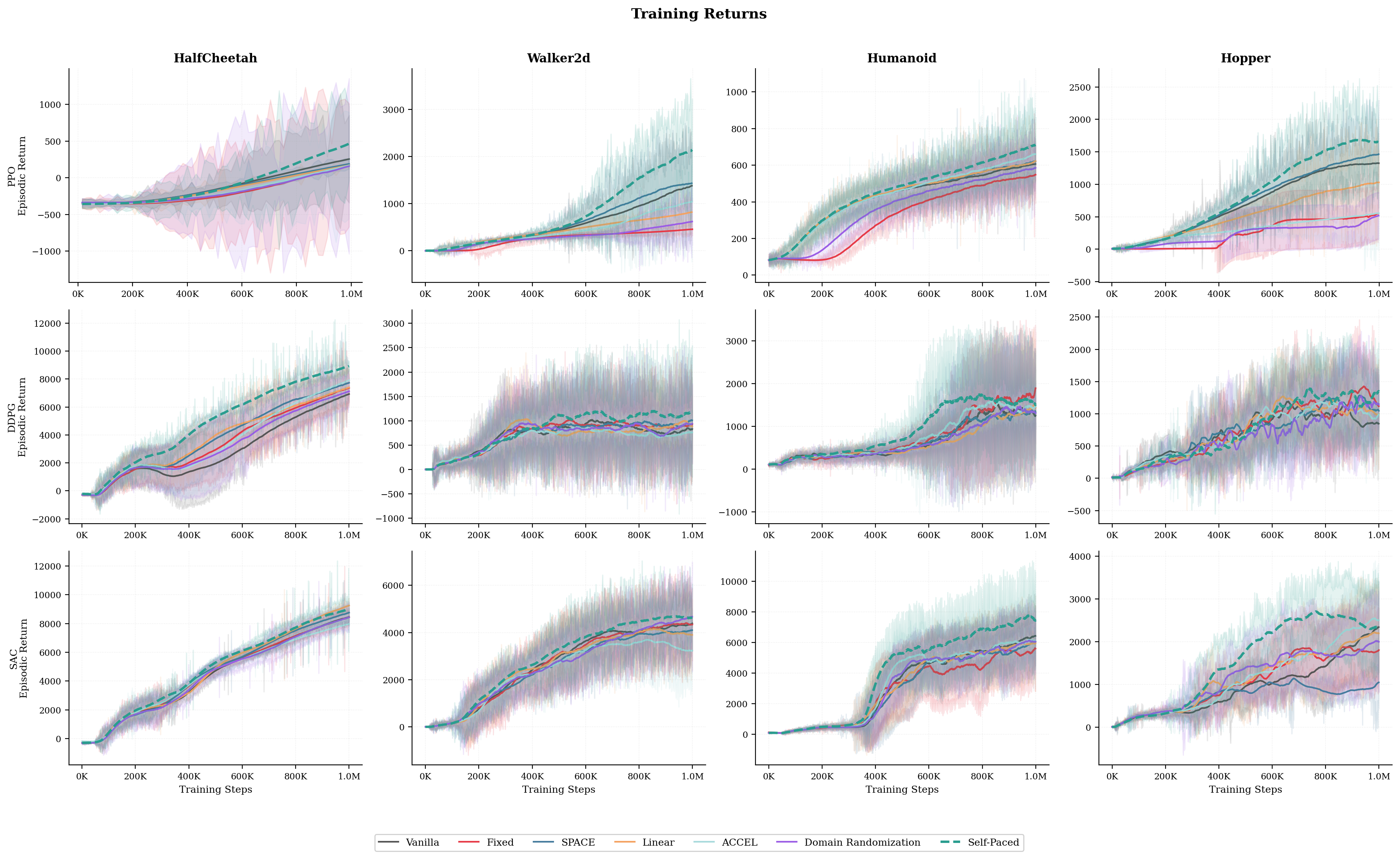}
    \caption{\textbf{Training curves on MuJoCo benchmarks.} We compare DR-SPCRL (labeled \textit{Self-Paced}) against six baselines across PPO, DDPG, and SAC algorithms. The solid and dashed lines represent the mean episodic return over 5 random seeds, and the shaded regions indicate the 95\% confidence interval. DR-SPCRL consistently achieves higher asymptotic performance and faster convergence compared to fixed and heuristic curriculum baselines, avoiding the failure modes observed in non-curriculum and heuristic robust methods.}
    \label{fig:training_curves}
\end{figure*}

\paragraph{Improved Training Performance via Adaptive Robustness.} 
As illustrated in Figure~\ref{fig:training_curves}, DR-SPCRL (represented by the dashed teal line) consistently outperforms the Fixed and Linear robust baselines and frequently exceeds the performance of the Vanilla non-robust agent. We attribute this performance gain to the regularization effect of the distributionally robust objective, since worst-case perturbations used to estimate the robust value function implicitly introduce diversity into the transition dynamics. This stochasticity enhances exploration, preventing the policy from overfitting to the deterministic nominal dynamics and getting stuck in local optima. Crucially, unlike the Fixed budget approach, which often results in flat learning curves (e.g., PPO on Hopper) by exposing the agent to insurmountable adversarial perturbations too early, DR-SPCRL maintains training stability. By dynamically scaling the uncertainty set size $\epsilon$ only when the dual variable $\beta^*$ indicates sufficient agent performance, our method provides a more effective and consistent learning signal throughout the training process.

\begin{figure*}[h]
    \centering
    \begin{tabular}{cc}
        \includegraphics[width=0.48\textwidth]{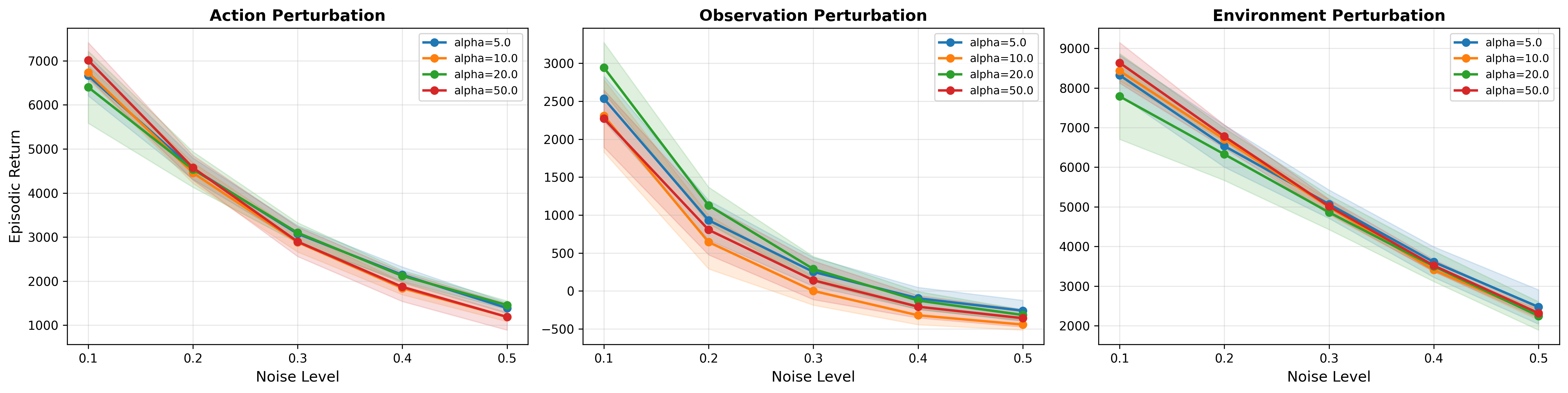} &
        \includegraphics[width=0.48\textwidth]{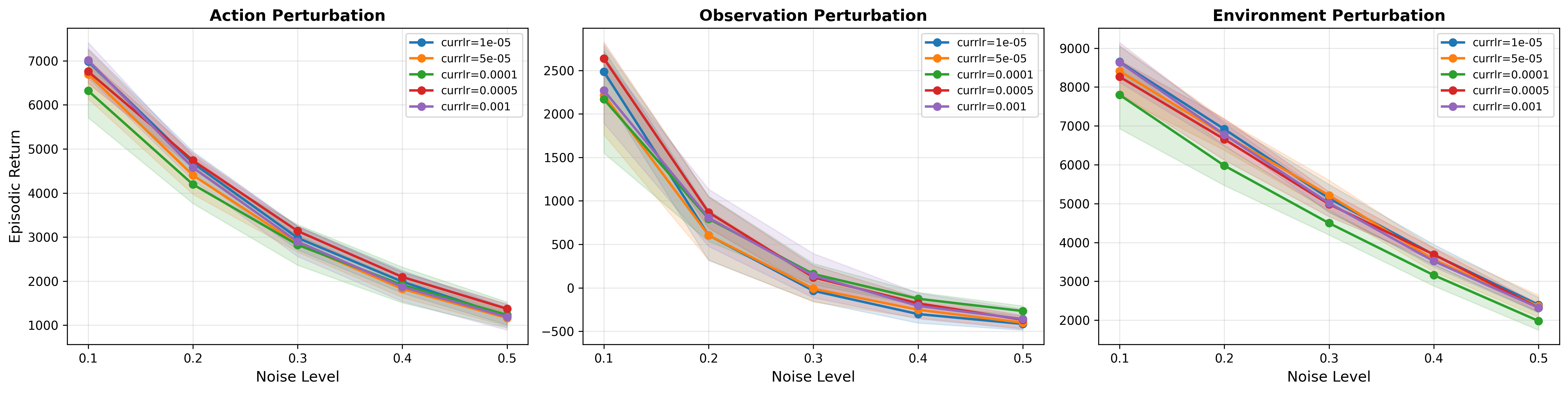} \\
        (a) DDPG: Varying pacing parameters $\alpha$ & (b) DDPG: Varying curriculum learning rates \\[1em]
        
        \includegraphics[width=0.48\textwidth]{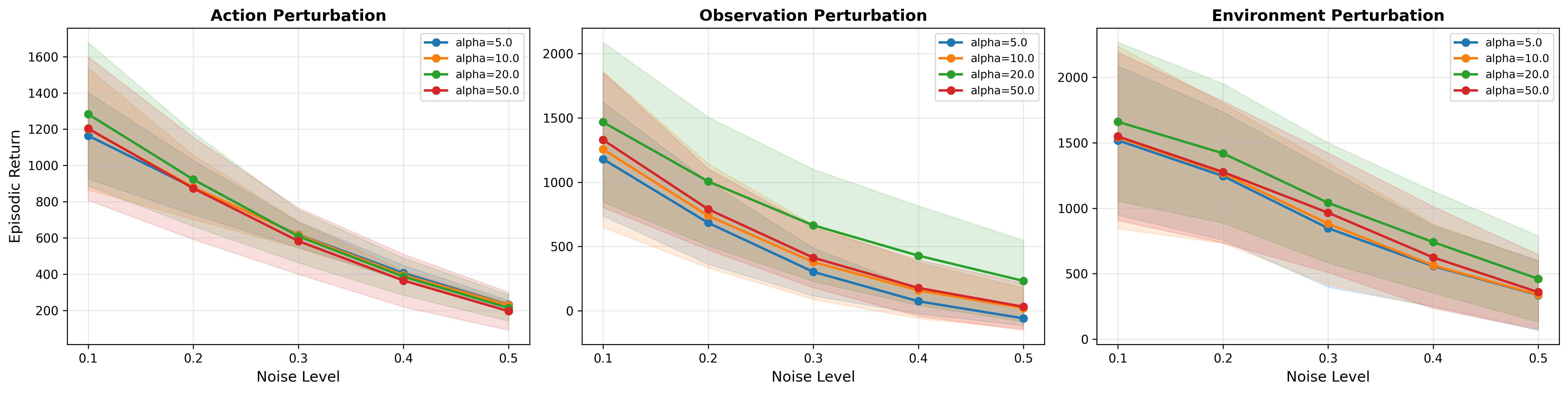} &
        \includegraphics[width=0.48\textwidth]{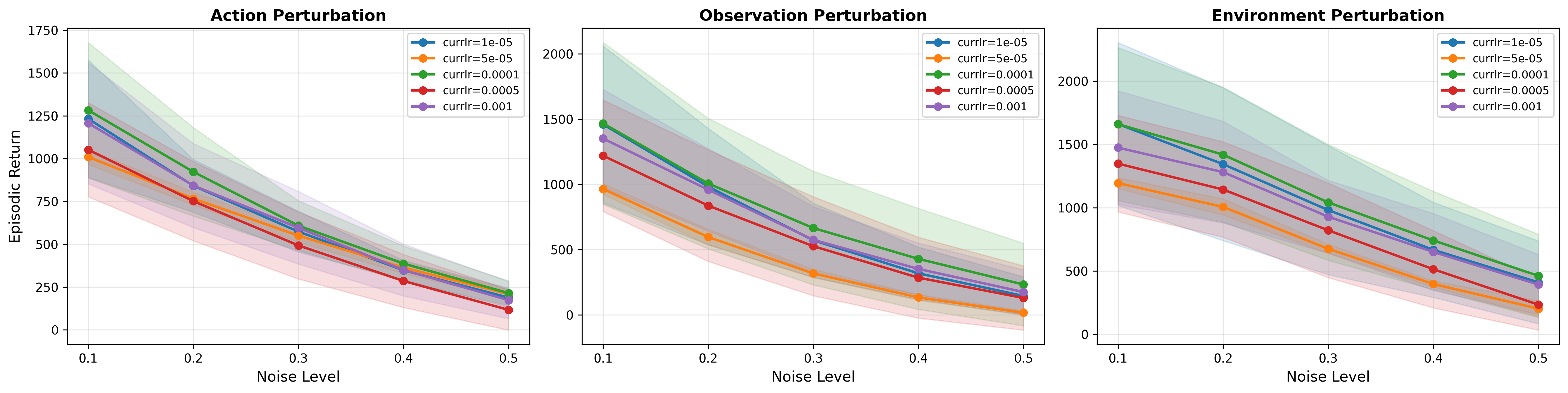} \\
        (c) PPO: Varying pacing parameters $\alpha$ & (d) PPO: Varying curriculum learning rates \\[1em]
        
        \includegraphics[width=0.48\textwidth]{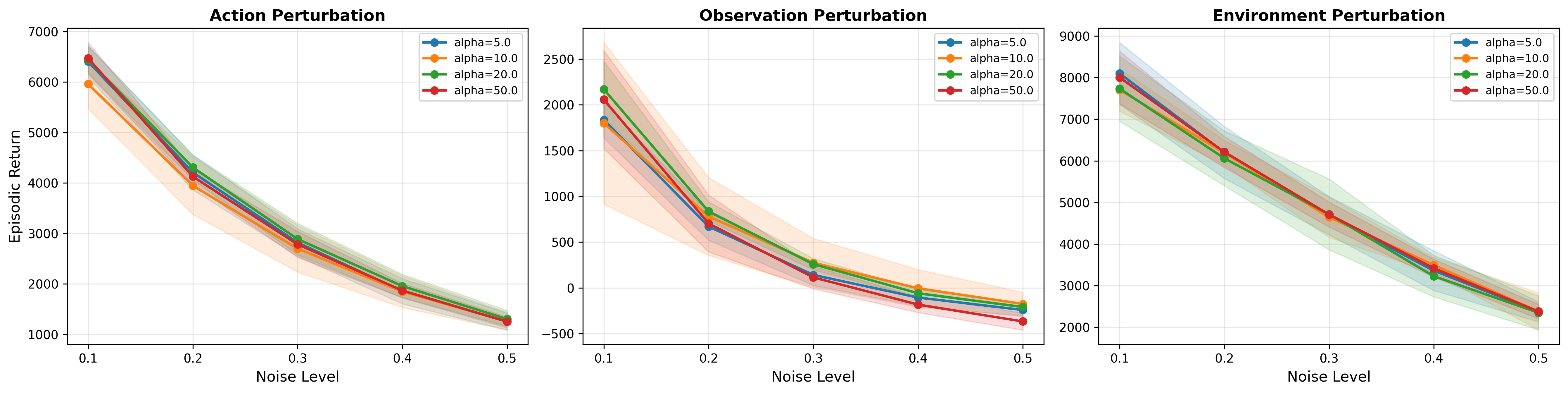} &
        \includegraphics[width=0.48\textwidth]{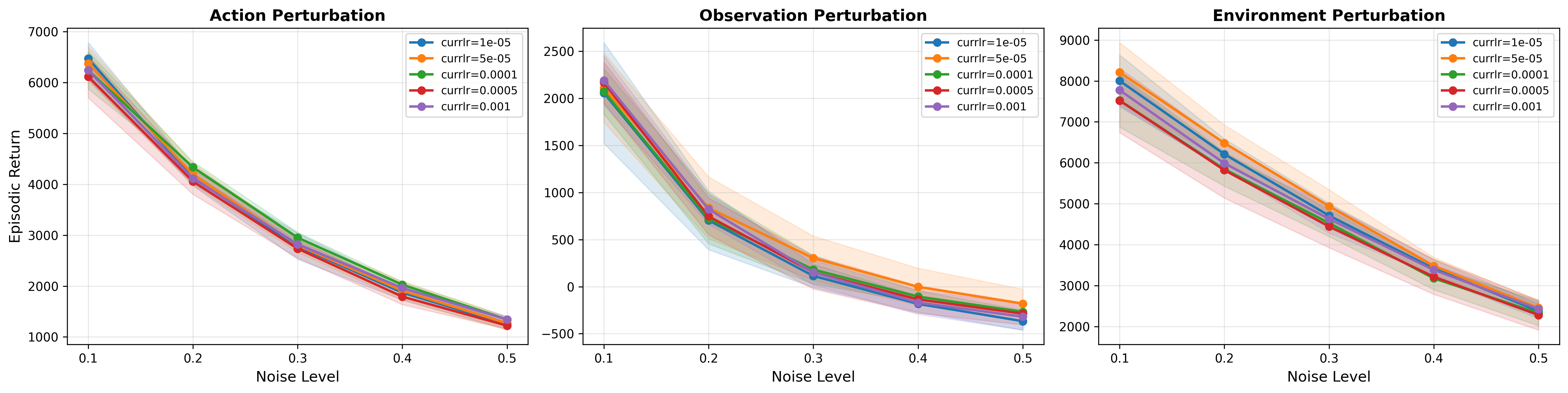} \\
        (e) SAC: Varying pacing parameters $\alpha$ & (f) SAC: Varying curriculum learning rates \\
    \end{tabular}
    \caption{Ablation studies evaluating DR-SPCRL’s robustness under fixed hyperparameters. Each panel shows episodic returns under observation, action, and environmental perturbations. Left column: results with fixed pacing parameters $\alpha$. Right column: results with fixed curriculum learning rates. Across all algorithms (DDPG, PPO, SAC), DR-SPCRL maintains high robustness and demonstrates stability despite suboptimal hyperparameter choices.}
    \label{fig:ablation_studies}
\end{figure*}

\section{Conclusion and Future Work}
In this paper, we introduce DR-SPCRL, a novel curriculum learning approach for distributionally robust reinforcement learning. Our method applies self-paced curriculum learning by treating the \(\epsilon\)-radius of the uncertainty set as the context for each training iteration. Experimental results across three baseline RL algorithms, 5 CRL algorithms, and four environments demonstrate that DR-SPCRL improves robustness and outperforms existing curriculum reinforcement learning methods.

We use the KL divergence to define the uncertainty set due to its tractable dual formulation, but our method can be extended to any $f$-divergence by the Legendre-Fenchel transform as long as a suitable estimation of the dual variable can be made. Future work can extend our curriculum to more diverse uncertainty set formulations, such as $\chi^2$, TV, and Wasserstein sets. Although DR-SPCRL is designed for single-agent settings, there is growing interest in multi-agent distributionally robust algorithms. Our framework could be extended to the multi-agent setting by employing a joint robust value function and appropriately adapting the formulation and algorithm. Future work could also explore integrating DR-SPCRL with model-based RL or planning methods, potentially allowing the agent to anticipate perturbations more efficiently and further improve robustness.


\bibliography{references}
\bibliographystyle{unsrt}

\appendix
\section{Theoretical Analysis}
\label{appendix: theoretical analysis}
\begin{theorem}[Finite-Time Convergence and Performance Bounds]
\label{thm: convergence}
Suppose the joint objective $F(\theta,\epsilon)$ is $L$-smooth and the stochastic gradient estimates are unbiased with variance bounded by $\sigma^2$. Further assume that the gradient of the robust value function, $\nabla J(\pi_\theta, \epsilon)$, is $L_\beta$-Lipschitz continuous with respect to $\epsilon$. Let $C_\gamma = \frac{\gamma}{1-\gamma}$ be the effective horizon scaling factor.
Let Algorithm~\ref{alg:dr-spcl} be run for $T \ge 4L^2$ iterations with step size $\eta=T^{-1/2}$. Then, the last iterate $(\pi_T, \epsilon_T)$ satisfies the following bounds:

\begin{enumerate}
    \item \textbf{Policy Stationarity at budget:} The policy $\pi_T$ is an approximate stationary point of the optimal robust objective $J(\cdot, \epsilon_{\mathrm{budget}})$, with expected gradient norm bounded by:
    \[
    \mathbb{E}\|\nabla_\theta J(\pi_T, \epsilon_{\mathrm{budget}})\|
    \;\le\; \mathcal{O}(T^{-1/4}) \;+\; \frac{L_\beta\,C_\gamma\,\mathbb{E}[\beta_T^*]}{2\alpha}.
    \]

    \item \textbf{Curriculum Approximation Gap:} The expected discrepancy between the robust value at \(\epsilon_{\mathrm{budget}}\) and \(\epsilon_T\) is bounded by:
    \[
    \mathbb{E}\big| J(\pi_T, \epsilon_{\mathrm{budget}}) - J(\pi_T, \epsilon_T) \big|
    \;\le\; \mathcal{O}(T^{-1/2}) \;+\; \frac{C_\gamma^2(\mathbb{E}[\beta_T^*])^2}{2\alpha}\left(1 + \frac{L_\beta}{2\alpha}\right).
    \]
\end{enumerate}
Here, $\beta_T^*$ denotes the expected optimal dual variable for the uncertainty set at $(\pi_T, \epsilon_T)$, and the constants hidden in $\mathcal{O}(\cdot)$ depend on $L$, $\sigma^2$, and the initial function value gap $F(z_0) - F_{\inf}$.
\end{theorem}

\begin{proof}
We analyze Algorithm~\ref{alg:dr-spcl} as stochastic gradient ascent on the penalized objective
\begin{align}
  F(\theta,\epsilon)
  &\;=\; J(\pi_\theta,\epsilon) - \alpha(\epsilon-\epsilon_{\mathrm{budget}})^2,
\end{align}
where $\alpha>0$ is a fixed penalty coefficient. Write $z_t=(\theta_t,\epsilon_t)$,
and let $F_{\inf}=\inf_z F(z)>-\infty$.
The stochastic gradient $g_t$ satisfies
\begin{align}
  \mathbb{E}[g_t\mid z_t] &= \nabla F(z_t), \label{eq:unbiased}\\
  \mathbb{E}[\|g_t-\nabla F(z_t)\|^2\mid z_t] &\le \sigma^2. \label{eq:variance}
\end{align}
The update is $z_{t+1}=z_t+\eta g_t$ with step size $\eta=T^{-1/2}$. Since $F$ is $L$-smooth, the ascent lemma gives
\begin{align}
  F(z_{t+1})
  &\;\ge\; F(z_t)+\langle\nabla F(z_t),z_{t+1}-z_t\rangle
          -\tfrac{L}{2}\|z_{t+1}-z_t\|^2 \\
  &\;=\; F(z_t)+\eta\langle\nabla F(z_t),g_t\rangle
         -\tfrac{L\eta^2}{2}\|g_t\|^2.
\end{align}
Taking conditional expectation $\mathbb{E}_t[\,\cdot\,]=\mathbb{E}[\,\cdot\mid z_t]$
and using \eqref{eq:unbiased}--\eqref{eq:variance},
\begin{align}
  \mathbb{E}_t[F(z_{t+1})]
  &\;\ge\; F(z_t)+\eta\|\nabla F(z_t)\|^2
           -\tfrac{L\eta^2}{2}\bigl(\|\nabla F(z_t)\|^2+\sigma^2\bigr) \\
  &\;=\; F(z_t)+\eta\Bigl(1-\tfrac{L\eta}{2}\Bigr)\|\nabla F(z_t)\|^2
         -\tfrac{L\eta^2\sigma^2}{2}.
\end{align}
For $T\ge 4L^2$ we have $\eta\le(2L)^{-1}$, so $1-L\eta/2\ge1/2$, and therefore
\begin{align}
  \mathbb{E}_t[F(z_{t+1})]
  &\;\ge\; F(z_t)+\tfrac{\eta}{2}\|\nabla F(z_t)\|^2-\tfrac{L\eta^2\sigma^2}{2}.
\end{align}
Rearranging and summing over $t=0,\ldots,T-1$,
\begin{align}
  \frac{\eta}{2}\sum_{t=0}^{T-1}\mathbb{E}\|\nabla F(z_t)\|^2
  &\;\le\; \mathbb{E}[F(z_0)-F(z_T)]+\frac{LT\eta^2\sigma^2}{2} \\
  &\;\le\; \bigl(F(z_0)-F_{\inf}\bigr)+\frac{LT\eta^2\sigma^2}{2},
\end{align}
where the second line uses $F(z_T)\ge F_{\inf}$.
Dividing by $T\eta/2$ and substituting $\eta=T^{-1/2}$:
\begin{align}
  \frac{1}{T}\sum_{t=0}^{T-1}\mathbb{E}\|\nabla F(z_t)\|^2
  &\;\le\; \frac{2(F(z_0)-F_{\inf})+L\sigma^2}{\sqrt{T}}
  \;\eqqcolon\; \frac{C_0}{\sqrt{T}}.
\end{align}
Let $\tau$ be drawn uniformly from $\{0,\ldots,T-1\}$. Then
\begin{align}
  \mathbb{E}\|\nabla F(z_\tau)\|^2 \;\le\; \frac{C_0}{\sqrt{T}}.
  \label{eq:stat}
\end{align}
Write $\xi_\theta=\nabla_\theta F(z_\tau)$ and $\xi_\epsilon=\nabla_\epsilon F(z_\tau)$,
so that $\|\xi_\theta\|^2+\xi_\epsilon^2=\|\nabla F(z_\tau)\|^2\le C_0/\sqrt{T}$
in expectation. Direct computation gives
\begin{align}
  \nabla_\epsilon F(\theta,\epsilon)
  = \nabla_\epsilon J(\pi_\theta,\epsilon)-2\alpha(\epsilon-\epsilon_{\mathrm{budget}}).
\end{align}
By the envelope theorem applied to the inner optimization defining
$J(\pi_\theta,\epsilon)$ (accounting for the infinite horizon sum of dual variables),
\begin{align}
  \nabla_\epsilon J(\pi_\tau,\epsilon_\tau) = -C_\gamma \beta_\tau^*,
\end{align}
where $C_\gamma = \frac{\gamma}{1-\gamma}$ and $\beta_\tau^*\ge0$ is the expected dual variable associated with the budget constraint. Hence
\begin{align}
  \xi_\epsilon = -C_\gamma \beta_\tau^*-2\alpha(\epsilon_\tau-\epsilon_{\mathrm{budget}}),
\end{align}
which rearranges to
\begin{align}
  \epsilon_{\mathrm{budget}}-\epsilon_\tau
  = \frac{C_\gamma \beta_\tau^*+\xi_\epsilon}{2\alpha}.
  \label{eq:budget-dev}
\end{align}
At a stationary point ($\xi_\epsilon=0$) this gives $\epsilon_{\mathrm{budget}}-\epsilon_\tau = C_\gamma \beta_\tau^*/(2\alpha)$, which implies the curriculum gap scales with the effective horizon $C_\gamma$. Since $\nabla_\theta F=\nabla_\theta J$, the $\theta$-stationarity residual satisfies
\begin{align}
  \|\nabla_\theta J(\pi_\tau,\epsilon_\tau)\| = \|\xi_\theta\|.
\end{align}
By $L_\beta$-Lipschitz continuity of $\nabla_\theta J$ in $\epsilon$
and \eqref{eq:budget-dev},
\begin{align}
  \|\nabla_\theta J(\pi_\tau,\epsilon_{\mathrm{budget}})\|
  &\;\le\; \|\xi_\theta\|+L_\beta|\epsilon_{\mathrm{budget}}-\epsilon_\tau| \\
  &\;\le\; \|\xi_\theta\|+\frac{L_\beta|C_\gamma \beta_\tau^*+\xi_\epsilon|}{2\alpha}.
  \label{eq:grad-at-budget}
\end{align}
Then, \eqref{eq:grad-at-budget} certifies that
$\pi_\tau$ is a near-stationary point of $\theta\mapsto J(\pi_\theta,\epsilon_{\mathrm{budget}})$
with residual bounded in expectation by
\begin{align}
  \mathbb{E}\|\nabla_\theta J(\pi_\tau,\epsilon_{\mathrm{budget}})\|
  &\;\le\; \mathbb{E}\|\xi_\theta\|
           +\frac{L_\beta}{2\alpha}\mathbb{E}|C_\gamma \beta_\tau^*+\xi_\epsilon| \\
  &\;\le\; C_0^{1/2}T^{-1/4}
           +\frac{L_\beta}{2\alpha}
             \Bigl(C_\gamma \mathbb{E}[\beta^*]+C_0^{1/2}T^{-1/4}\Bigr) \\
  &\;\le\; \underbrace{\frac{L_\beta\,C_\gamma\,\mathbb{E}[\beta^*]}{2\alpha}}_{\text{bias}}
           \;+\; \underbrace{C_1 T^{-1/4}}_{\text{opt.\ error}},
  \label{eq:policy-stat}
\end{align}
where we used Jensen's inequality and $C_1>0$ absorbs constants. By $L_\beta$-smoothness of $J(\pi_\tau,\,\cdot\,)$:
\begin{align}
  J(\pi_\tau, \epsilon_{\mathrm{budget}}) - J(\pi_\tau, \epsilon_\tau)
  &\;\le\; |\nabla_\epsilon J(\pi_\tau,\epsilon_\tau)|
           \cdot|\epsilon_{\mathrm{budget}}-\epsilon_\tau|
           +\frac{L_\beta}{2}(\epsilon_{\mathrm{budget}}-\epsilon_\tau)^2 \\
  &\;=\; |C_\gamma \beta_\tau^*|\cdot\frac{|C_\gamma \beta_\tau^*+\xi_\epsilon|}{2\alpha}
         +\frac{L_\beta}{8\alpha^2}(C_\gamma \beta_\tau^*+\xi_\epsilon)^2,
\end{align}
where we substituted $\nabla_\epsilon J = -C_\gamma \beta_\tau^*$ and \eqref{eq:budget-dev}. Taking expectations and using $\mathbb{E}[\xi_\epsilon^2]\le C_0/\sqrt{T}$:
\begin{align}
  \mathbb{E}[J(\pi_\tau, \epsilon_{\mathrm{budget}}) - J(\pi_\tau, \epsilon_\tau)]
  &\;\le\; \frac{C_\gamma^2 (\mathbb{E}[\beta^*])^2}{2\alpha}
           +\frac{L_\beta C_\gamma^2 (\mathbb{E}[\beta^*])^2}{4\alpha^2}
           +\frac{C_2}{\sqrt{T}} \\
  &\;=\; \frac{C_\gamma^2 (\mathbb{E}[\beta^*])^2}{2\alpha}\left(1 + \frac{L_\beta}{2\alpha}\right) + \frac{C_2}{\sqrt{T}}.
  \label{eq:curric-gap}
\end{align}
Combining \eqref{eq:policy-stat} and \eqref{eq:curric-gap}, the algorithm outputs the last iterate $z_T$ satisfying the stated bounds. The bounds show the error scales quadratically with the effective horizon $C_\gamma$ and the marginal cost of robustness $\beta^*$.
\end{proof}
\section{Further Results}
\label{appendix: further results}
\begin{figure}[ht]
    \centering
    \includegraphics[width=\linewidth]{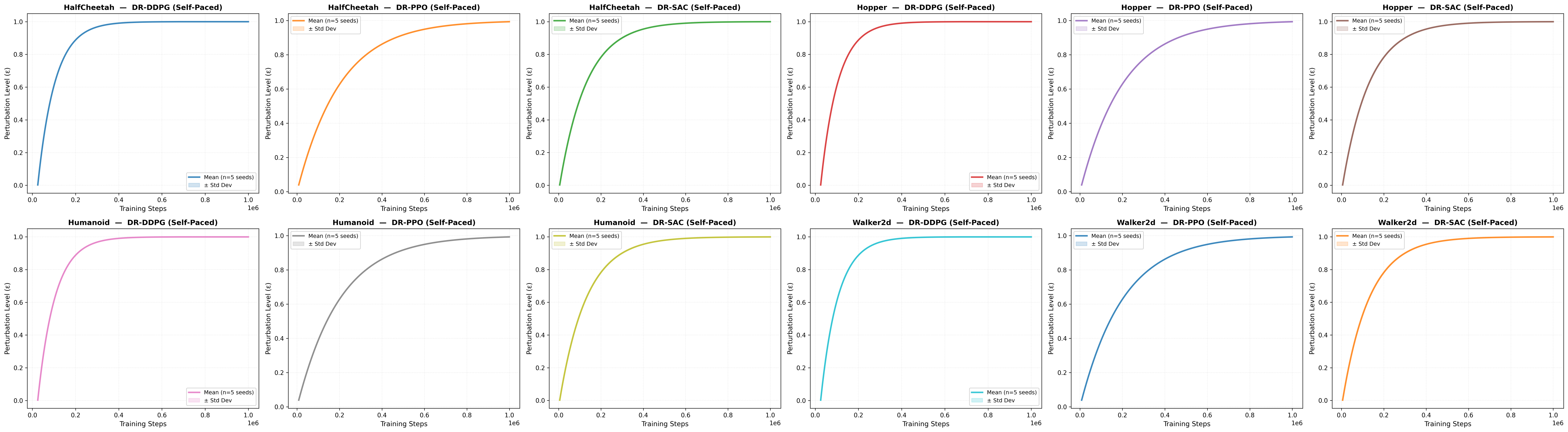}
    \caption{\textbf{Evolution of the robustness curriculum \(\epsilon_t\) during training across diverse environments and algorithms}. The plots illustrate the automatic curriculum scheduling performed by DR-SPCRL for DDPG, PPO, and SAC agents on HalfCheetah, Humanoid, Hopper, and Walker2d tasks. Starting from a nominal environment \(\epsilon = 0\), the algorithm progressively expands the uncertainty set radius towards the target budget \(\epsilon_{\mathrm{budget}} = 1\) based on the agent's learning progress and the dual variable \(\beta^*\). Solid lines represent the mean value of \(\epsilon_t\) over 5 random seeds, and shaded regions indicate the 95\% confidence interval.}
    \label{fig: epsilon curriculum plots}
\end{figure}

\paragraph{Evolution of the Robustness Curriculum.}
Figure~\ref{fig: epsilon curriculum plots} illustrates the adaptive scheduling of the robustness budget $\epsilon_t$ generated by DR-SPCRL across different environments and algorithms. The plots demonstrate that the algorithm successfully initiates training in the nominal environment ($\epsilon_t \approx 0$) to allow the agent to acquire basic competence, and subsequently expands the uncertainty set toward the target budget $\epsilon_{\text{budget}}=1$. A salient feature of these trajectories is the smoothness of the curriculum updates. Despite the inherent stochasticity of the dual variable estimates $\beta^*$ used to drive the curriculum, the resulting $\epsilon_t$ curves exhibit stable, monotonic growth with minimal oscillation. This smoothness indicates that the trust-region constraint and the pacing parameter $\alpha$ effectively filter high-variance gradients, allowing the agent to steadily assimilate robustness requirements without facing erratic jumps in difficulty that could destabilize the policy updates.

\begin{figure}
    \centering
    \includegraphics[width=\linewidth]{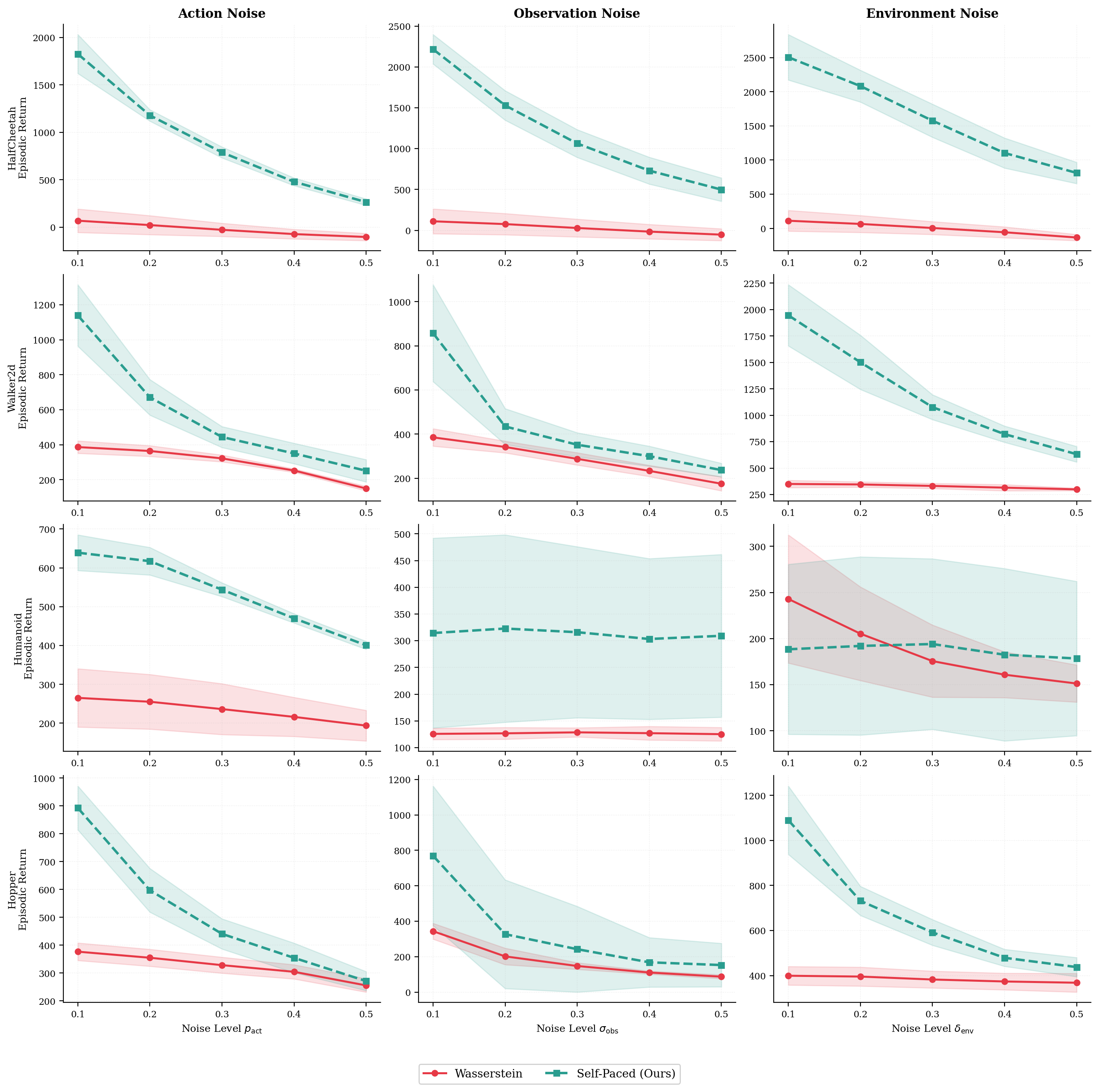}
    \caption{We benchmark DR-SPCRL (Self-Paced) against the Wasserstein Robust RL (WR2L) algorithm~\citep{abdullah2019wasserstein} to evaluate the benefits of our curriculum over model-based techniques. The plots show episodic returns under increasing Action Noise ($p_{\text{act}}$), Observation Noise ($\sigma_{\text{obs}}$), and Environment Parameter shifts ($\delta_{\text{env}}$). DR-SPCRL significantly outperforms WR2L, which suffers from excessive conservatism.}
    \label{fig:wasserstein comparison}
\end{figure}

\paragraph{Comparison with model-based Wasserstein Robust RL.}
We include the comparison with Wasserstein Robust RL (WR2L)~\citep{abdullah2019wasserstein} to explicitly contrast our adaptive curriculum approach against a leading distributionally robust baseline that optimizes for worst-case model deviations without a curriculum. W2RL learns an adversarial environmental parameterization based on the Wasserstein uncertainty set, and updates the parameter via a trust region using the dual formulation for the wasserstein divergence. As shown in Figure~\ref{fig:wasserstein comparison}, WR2L (red line) struggles to learn meaningful policies in higher-dimensional environments like HalfCheetah and Walker2d, resulting in near-zero returns across all perturbation levels. 
\end{document}